%% file: main.tex
\documentclass[11pt, a4paper, logo, copyright]{googledeepmind}

\usepackage[authoryear, sort&compress, round]{natbib}
\usepackage{multirow}
\usepackage{tcolorbox}
\usepackage{graphicx} 
\usepackage{subcaption}
\usepackage{float}
\usepackage{cleveref}
\usepackage{pifont}
\newcommand{\cmark}{\ding{51}}

\usepackage{textcomp}
\usepackage{afterpage}
\hypersetup{
    colorlinks=true,
}

\title{Capabilities of Gemini Models in Medicine}

\correspondingauthor{$\{$ksaab, taotu, alankarthi, natviv$\}$@google.com}

\reportnumber{} 

\renewcommand{\today}{2024-04-29}

\author[$\circ$,1]{Khaled Saab}
\author[$\circ$,$\ddagger$,1]{Tao Tu}
\author[$\circ$,1]{Wei-Hung Weng}
\author[$\circ$,2]{Ryutaro Tanno}
\author[*,2]{David Stutz}
\author[*,1]{Ellery Wulczyn}
\author[*,1]{\\Fan Zhang}
\author[*,1]{Tim Strother}
\author[*,1]{Chunjong Park}
\author[*,1]{Elahe Vedadi}
\author[*,1]{Juanma Zambrano Chaves}
\author[*,1]{\\Szu-Yeu Hu}
\author[*,1]{Mike Schaekermann}
\author[*,2]{Aishwarya Kamath}
\author[*,2]{Yong Cheng}
\author[*,2]{David G.T. Barrett}
\author[*,1]{\\Cathy Cheung}
\author[*,2]{Basil Mustafa}
\author[*,1]{Anil Palepu}
\author[*,1]{Daniel McDuff}
\author[*,2]{Le Hou}
\author[*,4]{Tomer Golany}
\author[*,1]{\\Luyang Liu}
\author[*,2]{Jean-baptiste Alayrac}
\author[*,2]{Neil Houlsby}
\author[*,2]{Nenad Tomasev}
\author[*,1]{Jan Freyberg}
\author[1]{Charles Lau}
\author[1]{Jonas Kemp}
\author[1]{Jeremy Lai}
\author[2]{Shekoofeh Azizi}
\author[1]{Kimberly Kanada}
\author[1]{SiWai Man}
\author[1]{Kavita Kulkarni}
\author[3]{Ruoxi Sun}
\author[2]{Siamak Shakeri}
\author[2]{Luheng He}
\author[2]{Ben Caine}
\author[2]{Albert Webson}
\author[2]{Natasha Latysheva}
\author[2]{Melvin Johnson}
\author[1]{\\Philip Mansfield}
\author[1]{Jian Lu}
\author[4]{Ehud Rivlin}
\author[1]{Jesper Anderson}
\author[1]{Bradley Green}
\author[1]{Renee Wong}
\author[1]{\\Jonathan Krause}
\author[2]{Jonathon Shlens}
\author[1]{Ewa Dominowska}
\author[2]{S. M. Ali Eslami}
\author[2]{Katherine Chou}
\author[2]{Claire Cui}
\author[2]{\\Oriol Vinyals}
\author[2]{Koray Kavukcuoglu}
\author[1]{James Manyika}
\author[1,2]{Jeff Dean}
\author[2]{Demis Hassabis}
\author[1]{Yossi Matias}
\author[1]{\\Dale Webster}
\author[2]{Joelle Barral}
\author[1]{Greg Corrado}
\author[1]{Christopher Semturs}
\author[*,2]{S. Sara Mahdavi}
\author[*,3]{Juraj Gottweis}
\author[*,1]{\\Alan Karthikesalingam}
\author[$\dagger$,1]{Vivek Natarajan}

\affil[$\circ$]{Co-first}
\affil[*]{Core}
\affil[$\ddagger$]{Technical Lead}
\affil[$\dagger$]{Senior Lead}
\affil[1]{Google Research}
\affil[2]{Google DeepMind}
\affil[3]{Google Cloud}
\affil[4]{Verily}

\begin{abstract}
Excellence in a wide variety of medical applications poses considerable challenges for AI, requiring advanced reasoning, access to up-to-date medical knowledge and understanding of complex multimodal data. Gemini models, with their strong general capabilities in multimodal and long-context reasoning, offer exciting possibilities in medicine. Building on these core strengths of Gemini 1.0 and Gemini 1.5, we introduce \textit{Med-Gemini}, a family of highly capable multimodal models that are specialized in medicine with the ability to seamlessly integrate the use of web search, and that can be efficiently tailored to novel modalities using custom encoders. We evaluate Med-Gemini on 14 medical benchmarks spanning text, multimodal and long-context applications, establishing new state-of-the-art (SoTA) performance on 10 of them, and surpass the GPT-4 model family on every benchmark where a direct comparison is viable, often by a wide margin. On the popular MedQA (USMLE) benchmark, our best-performing Med-Gemini model achieves SoTA performance of 91.1\% accuracy, using a novel uncertainty-guided search strategy, outperforming our prior best Med-PaLM 2 by 4.6\%. Our search-based strategy generalizes with SoTA performance on complex diagnostic challenges from the New England Journal of Medicine (NEJM) and the GeneTuring benchmark. On 7 multimodal benchmarks including NEJM Image Challenges and MMMU (health \& medicine), Med-Gemini improves over GPT-4V by an average relative margin of 44.5\%. We demonstrate the effectiveness of Med-Gemini's long-context capabilities through SoTA performance on a needle-in-a-haystack retrieval task from long de-identified health records and medical video question answering, surpassing prior bespoke methods using only in-context learning. Finally, Med-Gemini's performance suggests real-world utility by surpassing human experts on tasks such as medical text summarization and referral letter generation, alongside demonstrations of promising potential for multimodal medical dialogue, medical research and education. Taken together, our results offer compelling evidence for the promise of Med-Gemini in many areas of medicine, although further rigorous evaluation will be crucial before real-world deployment in this safety-critical domain.
\end{abstract}

\begin{document}

\maketitle

\clearpage
\begin{figure}[ht!]
    \centering
    \vspace{-0.6cm}
    \includegraphics[width=\textwidth,keepaspectratio,height=0.8\textheight]{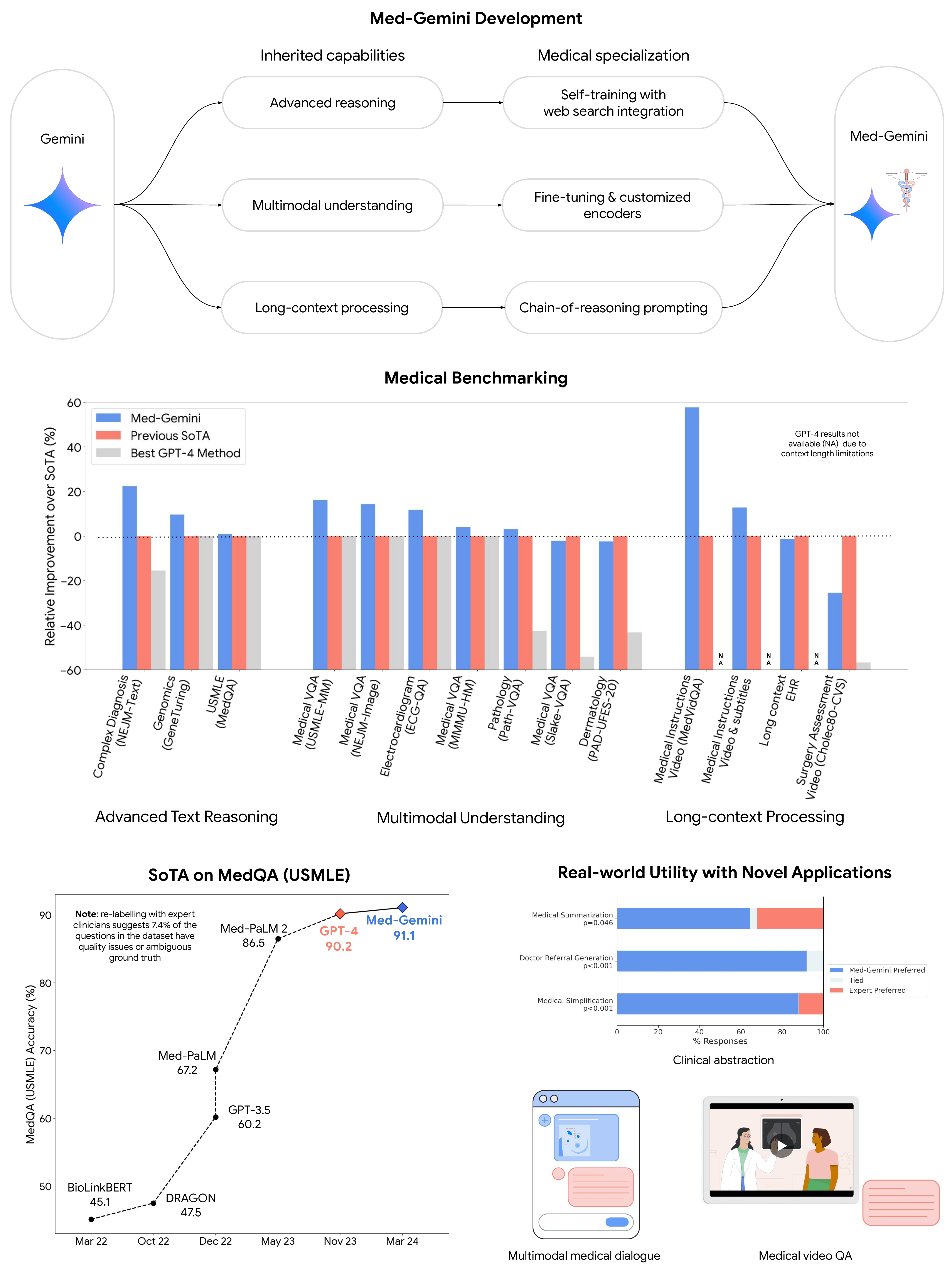}
    \vspace{-0.1cm}
    \caption{ 
    \footnotesize
    \textbf{Overview of our contributions.} We introduce \textit{Med-Gemini}, a family of highly capable, multimodal medical models built upon Gemini. We enhance our models' clinical reasoning capabilities through self-training and web search integration, while improving multimodal performance via fine-tuning and customized encoders. Med-Gemini models achieve state-of-the-art (SoTA) performance on 10 out of 14 medical benchmarks that span text, multimodal, and long-context applications, and surpass the GPT-4 model family on every benchmark where a direct comparison could be made.
    The bar chart shows the relative percentage gains from our models over prior SoTA across the benchmarks. In particular, on the MedQA (USMLE) benchmark, we attain a new SoTA surpassing our prior best (Med-PaLM 2) by a significant margin of 4.6\%. Moreover, re-annotation of the dataset with expert clinicians reveals that 7.4\% of questions are deemed unfit for evaluation as they either lack key information, have incorrect answers, or support multiple plausible interpretations. We account for these data quality issues to characterize more precisely the performance of our model. Med-Gemini models excel in multimodal and long-context capabilities as evidenced by their SoTA performance on several benchmarks including needle-in-a-haystack retrieval from long, de-identified health records, and medical video question answering benchmarks. Moving beyond benchmarks, we also demonstrate the real-world potential of Med-Gemini through quantitative evaluation on medical summarization, referral letter generation, and medical simplification tasks where our models outperform human experts, in addition to qualitative examples of multimodal medical dialogue.
    \vspace{-1.5cm}
    }
    \label{fig:overview}
\end{figure}

\input{introduction}
\input{methods}
\input{evaluation}
\input{results}
\clearpage

\input{discussion}

\section{Acknowledgements}
This project was an extensive collaboration between many teams at Google Research and Google DeepMind.
We thank Taylan Cemgil, Jake Sunshine, Daniel Golden, Pete Clardy, Zoubin Ghahramani and Dr. Gary Peltz (Stanford University) for their comprehensive review and detailed feedback on the manuscript. We also thank Sami Lachgar, Lauren Winer, John Guilyard, and Maggie Shiels for contributions to the narratives and visuals. We thank Yun Liu for discussions, design, and preliminary analysis for the MedQA label uncertainty experiments. We are grateful to Noam Velan, Ira Ktena, Eric Aboussouan, Karan Singhal, Shashir Reddy, Aza Tulepbergenov, Priya Gupta, Rory Sayres, Naama Hammel, Jen McKay, Peter Clardy, Chu-ling Ko, Abhinav Das, Haiyang Yu, Chang Liu, Yuchen Liu, Erica Moreira, Jordan Grimstad, Brett Hatfield, Gordon Turner, Jackie Barr, Jim Winkens, Jackie Barr, Brian Cappy, Pinal Bavishi, Tim McConnell, Ines Mezzorag, Annisah Um’rani, Christian Wright, Divya Pandya, Daireen Garcia, Prachant Bradwell, Alyssa Pierce, Sarah-Jane Allen, Erica Harland, Jennifer Ye, Praney Mittal, Donny Cheung, Andy Crowne and Preeti Singh for their valuable technical support during our research. Finally, we are grateful to Shravya Shetty, Sushant Prakash, Susan Thomas, Michael Howell, Karen DeSalvo, and Zoubin Ghahramani for their support of this project.

\section{Data Availability}
Except for the three clinical abstraction tasks, the remaining datasets used for development, benchmarking and evaluation of the AI systems are open source or otherwise accessible publicly with permissions. We will make our re-annotation of the MedQA (USMLE) dataset publicly available.

\section{Code Availability}
We are not open-sourcing model code and
weights due to the safety implications of unmonitored use of such a system in medical settings. In the interest
of responsible innovation, we will be working with research partners, regulators, and providers to validate and
explore safe onward uses of our medical models and expect to make them available via Google Cloud APIs in due course.

\section{Competing Interests}
This study was funded by Alphabet Inc and/or a subsidiary thereof (`Alphabet'). All authors are (or were) employees of Alphabet and may own stock as part of the standard compensation package.

\clearpage
\bibliography{main}

\clearpage
\appendix
\input{appendix}

\end{document}

%% file: introduction.tex
\section{Introduction}

Medicine is a multifaceted endeavor. A clinician's day-to-day work involves patient consultations, where clear communication of diagnoses, treatment plans, and empathy are essential for building trust. Complex cases necessitate deeper understanding of the patient's history within the electronic medical record, along with multimodal reasoning from medical images and other diagnostics. To guide their decisions under uncertainty, clinicians must stay abreast of the latest medical information from a wide variety of authoritative sources that can range from research publications to procedural videos. The art of care delivery hinges on a clinician's ability to perform advanced clinical reasoning, synthesize complex information from diverse and multimodal sources, and collaborate effectively with other clinicians to help people in their care journeys. Although artificial intelligence (AI) systems can assist individual medical tasks~\citep{rajpurkar2022ai} and demonstrate early promise towards multimodal multi-task ``generalist'' medical uses~\citep{tu2024towardsmpm,moor2023foundation}, the development of more sophisticated reasoning, multimodal, and long-context understanding capabilities would enable significantly more intuitive and helpful assistive tools for clinicians and patients alike.

The advent of large language models (LLMs) and large multimodal models (LMMs), like GPT-4~\citep{achiam2023gpt}, PaLM~\citep{chowdhery2023palm} and Gemini~\citep{team2023gemini}, showed that such models effectively encode clinical knowledge and can perform impressively in medical question answering benchmarks, even for complex cases and scenarios requiring specialized knowledge~\citep{kanjee2023accuracy, eriksen2023use, antaki2023capabilities}. However, performance on such tasks is far from indicative of real-world utility. The unique nature of medical data and the critical need for safety demand specialized prompting~\citep{nori2023can}, fine-tuning, or potentially both along with careful alignment of these models~\citep{ouyang2022training}.

Medically fine-tuned LLMs~\citep{singhal2023large, luo2022biogpt, toma2023clinical} can also provide high-quality long-form answers to nuanced and open-ended medical questions asked by millions of internet users, with Med-PaLM 2 surpassing physicians on axes such as factuality, reasoning, harm, and bias~\citep{singhal2023towards}. The potential extends beyond question answering. LMMs~\citep{moor2023med, li2024llava} such as Flamingo-CXR and Med-PaLM M are comparable with radiologists in controlled settings for generating radiology reports~\citep{huang2023generative,tu2024towardsmpm,tanno2024consensus}. In the more challenging setting of text-based diagnostic consultations with patient actors, the Articulate Medical Intelligence Explorer (AMIE) model outperformed primary care physicians on several evaluation axes for diagnostic dialogue~\citep{tu2024towardsamie}.

Despite these promising results, there are considerable opportunities for improvement in performance. LLMs demonstrate suboptimal clinical reasoning under uncertainty, with confabulations and bias remaining key challenges~\citep{umapathi2023med, omiye2023large}. The use of tools and up-to-date medical information~\citep{zakka2024almanac} to accomplish medical tasks remains a challenge for LLMs, alongside effective collaboration with clinicians~\citep{mcduff2023towards}. Additionally, their ability to handle complex multimodal medical data (for example, integrating images, videos, and de-identified health records over time) is currently limited~\citep{tu2024towardsmpm}. Although these capabilities are particularly meaningful in medical applications, improvements in performance might be relevant beyond the medical domain. Tasks and benchmarks developed to measure and accelerate the progress of medical LLMs will be broadly impactful.

The Gemini models, as detailed in the Gemini 1.0 and 1.5 technical reports \citep{team2023gemini, team2024gemini}, are a new generation of highly capable multimodal models with novel foundational capabilities that have the potential to address some of these key challenges for medical AI. The models are transformer decoder models~\citep{vaswani2017attention, brown2020language} enhanced with innovations in architecture, optimization and training data, enabling them to exhibit strong capabilities across various modalities including images, audio, video, and text. The recent addition of the mixture-of-experts architecture~\citep{shazeer2017outrageously, fedus2022switch} allows the Gemini models to efficiently scale and reason over significantly longer and more complex data at inference time.

Building on the strengths of the Gemini models, we present \textit{Med-Gemini}, a family of models fine-tuned and specialized for medicine. The notion of generalist medical AI models has received considerable attention with impressive demonstrations of the possibilities for such systems~\citep{tu2024towardsmpm}. However, while the generalist approach is an meaningful research direction for medicine, real world considerations present trade-offs and requirements for task-specific optimizations which are at odds with each other. In this work, we do not attempt to build a generalist medical AI system. Rather, we introduce a family of models, each optimized for different capabilities and application-specific scenarios, considering factors such as training data, compute availability, and inference latency.  

Med-Gemini inherits Gemini's foundational capabilities in language and conversations, multimodal understanding, and long-context reasoning. For language-based tasks, we enhance the models' ability to use web search through self-training and introduce an inference time uncertainty-guided search strategy within an agent framework. This combination enables the model to provide more factually accurate, reliable, and nuanced results for complex clinical reasoning tasks. This leads to the state-of-the-art (SoTA) performance of 91.1\% accuracy on MedQA (USMLE)~\citep{jin2021disease} surpassing prior Med-PaLM 2 models by 4.6\%. We further conduct a careful examination of the MedQA (USMLE) data quality through relabelling with multiple independent expert clinicians, identifying unanswerable questions due to missing information and errors, enabling reliable analysis and characterization of our SoTA performance. The uncertainty-guided search strategy generalizes and leads to SoTA performance on the New England Journal of Medicine (NEJM) clinico-pathological conference (CPC) cases~\citep{kanjee2023accuracy, mcduff2023towards} and the GeneTuring benchmark~\citep{hou2023geneturing}. Beyond their strong performance on such benchmarks, our models suggest real-world utility by performing favorably when compared to human physicians on tasks such as medical note summarization and clinical referral letter generation.

As Gemini models are trained to accommodate textual input interleaved with a wide variety of other data modalities, they are known to excel in multimodal tasks. This confers impressive out-of-the-box SoTA performance on some multimodal medical benchmarks such as the NEJM Image Challenge. However, their performance can be further improved when dealing with specialized medical modalities not heavily represented in their pretraining data. We address this through multimodal fine-tuning and demonstrate the models' adaptability to novel medical modalities using customized encoders leading to SoTA performance on benchmarks such as Path-VQA~\citep{he2020pathvqa} and ECG-QA~\citep{oh2023ecgqa} among others. We qualitatively showcase our models' capabilities for clinically-meaningful multimodal conversation on a variety of both in-distribution and out-of-distribution data modalities.

Finally, the long-context capabilities of Gemini models open many exciting possibilities for application in medicine, given how frequently a clinically-meaningful decision requires parsing of large amounts of data with significant risks of ``information overload''~\citep{sbaffi2020information}. Our Med-Gemini models configured for long-context processing are able to seamlessly analyze complicated and long-form modalities such as de-identified electronic health records (EHRs) and videos. We demonstrate the effectiveness of these capabilities with impressive performance on the ``needle-in-a-haystack'' long EHR understanding~\citep{johnson2019mimic}, medical instructional video question answering~\citep{gupta-demner-fushman-2022-overview}, surgical action recognition from video~\citep{goodman2021real}, and the Critical View of Safety (CVS) assessment of surgical video~\citep{strasberg2010rationale} tasks. 

The advances of Med-Gemini have great promise, but it remains crucial to carefully consider the nuances of the medical field, acknowledge the role of AI systems as assistive tools for expert clinicians, and conduct rigorous validation before real-world deployments at scale.

Our key contributions are summarized below: 
\begin{itemize}
    \item \textbf{Med-Gemini}, our new family of multimodal medical models: We introduce a new family of highly capable multimodal medical models, built upon Gemini. Med-Gemini demonstrates important advancements in clinical reasoning, multimodal, and long-context capabilities. They are further fine-tuned to make use of web search for current information and can be customized to novel medical modalities through the use of modality-specific encoders.
    \item \textbf{Comprehensive benchmarking}: We evaluate Med-Gemini's capabilities on a suite of 25 tasks across 14 medical benchmarks, encompassing text, multimodal, and long-context applications. To the best of our knowledge, this is the most comprehensive benchmarking of multimodal medical models to date.
    \item \textbf{SoTA results on clinical language tasks}: Med-Gemini optimized for clinical reasoning reaches a SoTA performance of 91.1\% on MedQA (USMLE) using a novel uncertainty-guided search strategy. We quantify and characterize our performance improvements through a careful re-annotation of the MedQA dataset with clinical experts, finding these improvements to be meaningful. We further demonstrate the effectiveness of the search strategy through SoTA performance on NEJM CPC and GeneTuring benchmarks.
    \item \textbf{Multimodal and long-context capabilities}: Med-Gemini attains SoTA performance on 5 out of 7 multimodal medical benchmarks evaluated in this study. We demonstrate the effectiveness of multimodal medical fine-tuning and the ability to customize to novel medical modalities such as electrocardiograms (ECGs) using specialized encoder layers. Med-Gemini also exhibits strong long-context reasoning capabilities, attaining SoTA on challenging benchmarks such as ``needle-in-the-haystack'' tasks in lengthy electronic health records or benchmarks for medical video understanding. 
    In addition, in forthcoming work, we will also rigorously explore the capabilities of Gemini in radiology report generation.
    \item \textbf{Real-world utility of Med-Gemini}: Beyond performance on popular medical benchmarks, we preview the potential real-world utility of Med-Gemini through quantitative evaluations on tasks such as medical note summarization, clinical referral letter generation, and EHR question answering. We further showcase qualitative examples in multimodal diagnostic dialogues and applications of the models' long-context capabilities for medical education, clinician-facing tools, and biomedical research. We note that such uses (particularly in safety-critical areas like diagnosis) would require considerable further research and development.
\end{itemize}

%% file: methods.tex
\section{Methods}
As introduced in the Gemini technical reports~\citep{team2024gemini, team2023gemini}, the Gemini ecosystem encompasses a suite of models varying in size, modality encoders, and architectures, trained on a wide variety of high quality data across many modalities. 
The Gemini models exhibit state-of-the-art results across a diverse array of language, reasoning, coding, multilingual, image, and video benchmarks.
Notably, the Gemini 1.0 Ultra model excels in language-based tasks that require complex reasoning, and the Gemini 1.5 Pro model adds the ability to efficiently handle and make use of long-context inputs spanning millions of tokens and/or multimodal inputs such as hours of video or tens of hours of audio. Gemini 1.0 Nano is the smallest model variant in the Gemini model family that can run efficiently on-device. 

We develop our Med-Gemini models by building on the Gemini family, focusing on the following capabilities and methods:
\begin{enumerate}
    \item \textbf{Advanced reasoning via self-training and web search integration}: For language tasks that require less complex reasoning, such as summarizing medical notes and creating referral letters, we introduce \textit{Med-Gemini-M 1.0} by fine-tuning the Gemini 1.0 Pro model. For other tasks that require more advanced reasoning, we introduce \textit{Med-Gemini-L 1.0} by fine-tuning the Gemini 1.0 Ultra model using a self-training method to enable the models to efficiently use web search. We develop a novel uncertainty-guided search strategy at inference time to improve performance on complex clinical reasoning tasks. 
    \item \textbf{Multimodal understanding via fine-tuning and customized encoders}: The Gemini models are natively multimodal and have demonstrated impressive zero-shot performance on many multimodal benchmarks. However, the unique nature and heterogeneity of some medical modalities require fine-tuning to achieve the best possible performance. We introduce \textit{Med-Gemini-M 1.5} by performing fine-tuning with Gemini 1.5 Pro on a suite of multimodal medical datasets. We introduce \textit{Med-Gemini-S 1.0} and demonstrate the Gemini models' capability to adapt to novel medical modalities using specialized encoders with the Gemini 1.0 Nano model.
    \item \textbf{Long-context processing with chain-of-reasoning}: For the long-context processing tasks, we re-use \textit{Med-Gemini-M 1.5} with a long-context configuration. In addition, we also develop a novel inference-time chain-of-reasoning technique inspired by~\cite{tu2024towardsamie} to enable better understanding of long EHRs.
\end{enumerate}

\subsection{Advanced reasoning via self-training and web search integration}

Clinical reasoning is a fundamental skill that underpins successful care. Although it is a broad field with many definitions, clinical reasoning can be conceptualized as an iterative process by which a physician integrates their own clinical knowledge with initial patient information to form a case representation. This representation is then used to guide the iterative acquisition of additional information until a confidence threshold is reached to support a final diagnosis with plans for treatment and management \citep{gruppen2017clinical}. During this process, a physician may reason across many diverse inputs, such as patient symptoms, medical and socio-economic history, investigations and lab tests, prior responses to treatments and other wider factors such as epidemiological data. 
Moreover, many of these inputs have a time component, such as a series of evolving symptoms, lab measurements over time, or the various temporal data that is collected for monitoring health, such as electrocardiograms (ECGs). 
Medical knowledge is highly non-stationary, with reducing ``doubling times'' in the volume of medical information driven by the rapid pace of research~\citep{densen2011challenges,grandage2002less}. To ensure that their outputs reflect the latest information in this domain, LLMs might ideally not only possess strong reasoning capabilities but also be able to integrate up-to-date information, for example, from authoritative web sources. This grounding in external knowledge has the potential to reduce uncertainty in the model's responses, but requires an informed approach to information retrieval itself. The key challenge we aim to tackle with our medical fine-tuning of Gemini 1.0 Ultra is to improve the model's ability to make the most helpful web search queries and integrate their results in the reasoning process to generate accurate answers. The resulting model is Med-Gemini-L 1.0.

Instruction fine-tuning has been shown to improve the clinical reasoning ability of LLMs \citep{singhal2023towards}.
A prevalent instruction-tuning dataset is MedQA~\citep{jin2021disease}, which consists of multiple-choice questions representative of US Medical License Exam (USMLE) questions, that are designed to assess medical knowledge and reasoning across diverse scenarios with a large number of variables of interest~\citep{jin2021disease}. However, MedQA only provides a multiple-choice ground truth, and lacks expert demonstrations of the reasoning process necessary to train LLMs for clinical reasoning across diverse settings. As a result, LLMs fine-tuned on MedQA, such as Med-PaLM 2~\citep{singhal2023towards}, still exhibit significant reasoning shortcomings.
This, coupled with the lack of access to web search in such systems, results in factuality errors that compound in downstream reasoning steps or lead to models adopting premature conclusions without considering all possible reasoning pathways.

\paragraph{Fine-tuning datasets for language-based tasks} 
Collecting expert demonstrations of clinical reasoning, including how experts make informed use of knowledge retrieval tools such as web search, is both time-consuming and difficult to scale. To overcome this, we generate two novel datasets with self-training as described below: MedQA-R (Reasoning), which extends MedQA with synthetically generated reasoning explanations, or ``Chain-of-Thoughts'' (CoTs), and MedQA-RS (Reasoning and Search), which extends MedQA-R with instructions to use web search results as additional context to improve answer accuracy.

To add further variety to the fine-tuning data mixture of Med-Gemini-L 1.0, we also add a long-form question answering dataset, which consists of $260$ expert-crafted long-form responses to questions from HealthSearchQA, LiveQA, and MedicationQA in the MultiMedQA benchmark \citep{singhal2023large}, along with a medical summarization dataset, consisting of $65$ clinician-written summaries of medical notes from MIMIC-III \citep{johnson2016mimic}. We provide an overview of the datasets for language-based instruction fine-tuning datasets in Table \ref{tab:text_ft}.

\paragraph{Self-training with search}
Inspired by the recent success of self-training for synthetic data generation \citep{tu2024towardsamie}, we implement an iterative data-generation framework targeted at curating high-quality synthetic examples of clinical reasoning with web search use.

\begin{figure} 
    \centering
    \includegraphics[width=\textwidth]{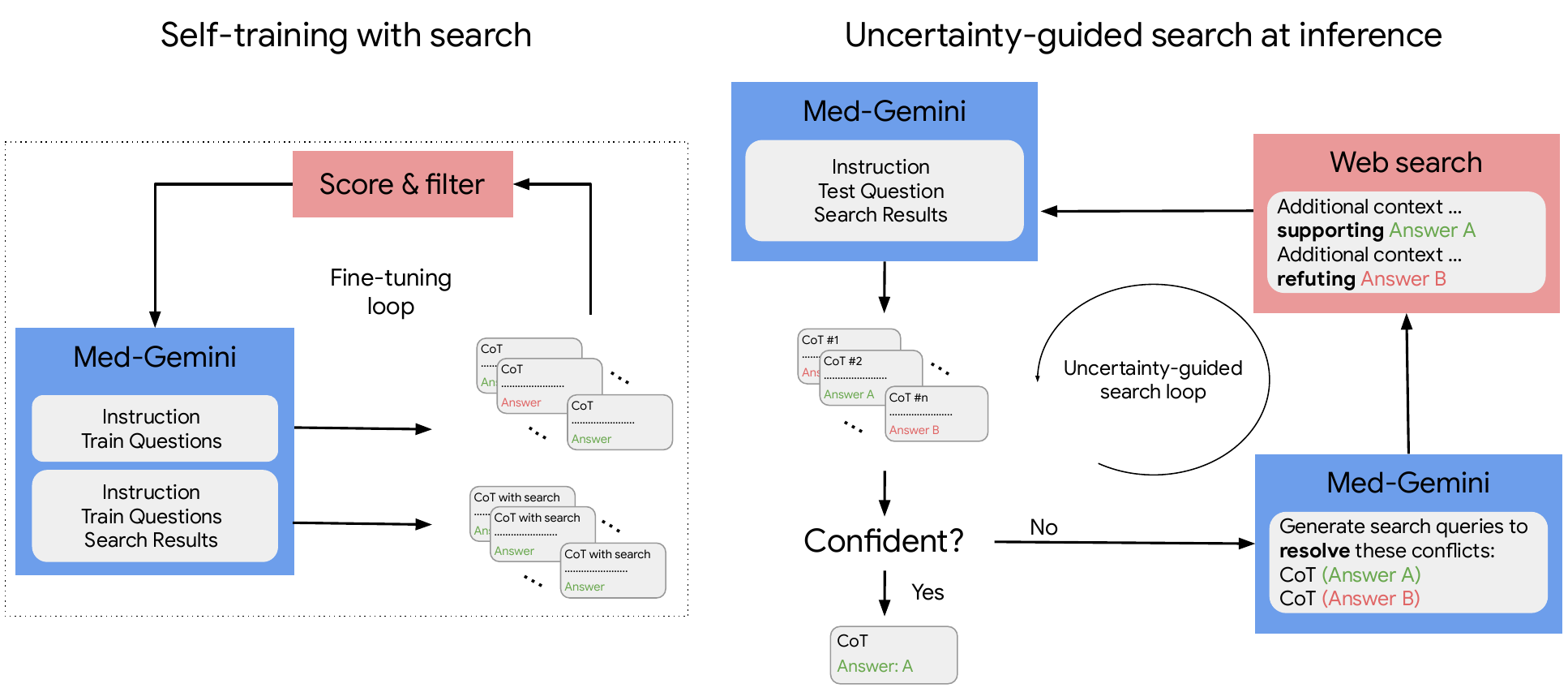}
    \caption{\footnotesize\textbf{Self-training and search tool-use.} The left panel illustrates the self-training with search framework used to fine-tune Med-Gemini-L 1.0 for advanced medical reasoning and use of web search. This framework iteratively generates reasoning responses (CoTs) with and without web search, improving the model's ability to utilize external information for accurate answers. The right panel illustrates Med-Gemini-L 1.0's uncertainty-guided search process at inference time. This iterative process involves generating multiple reasoning paths, filtering based on uncertainty, generating search queries to resolve ambiguity, and incorporating retrieved search results for more accurate responses.}
    \label{fig:search_diagram}
\end{figure}

As depicted in the left panel of~\Cref{fig:search_diagram}, we generate two reasoning paths, or CoTs, per training question: one without access to external information from search, and one that integrates search results as additional context during the CoT generation. Our self-training with search framework consists of the following key ingredients:
\begin{itemize}
    \item \textbf{Web search}: For each question, we prompt Med-Gemini-L 1.0 to generate search queries whose results would help answer the medical question. We then pass the search queries to a web search API and retrieve search results. 
    \item \textbf{In-context demonstrations}: For each type of reasoning response path, we hand-curate five expert demonstrations as seed with accurate clinical reasoning, explaining why the ground-truth answer is the best suited versus other potentially valid answers. For question examples with search results, the demonstrations explicitly refer to, and quote, the helpful information in the search results to best answer the question.
    \item \textbf{Generating CoTs}: We prompt Med-Gemini-L 1.0 to generate CoTs using the in-context seed demonstrations over the train set. Before fine-tuning the model on the generated CoTs, we filter out the ones that lead to erroneous predictions.
    \item \textbf{Fine-tuning loop}: After fine-tuning Med-Gemini-L 1.0 on the generated CoTs, the model's ability to follow the reasoning style and search integration of expert demonstrations improves. We then use the improved model to re-generate the CoTs, and iteratively repeat this self-training process until the model's performance saturates.
\end{itemize}
Below we provide a MedQA-RS example of an input prompt, along with the retrieved search results and an example of a generated CoT, which is then used to further fine-tune Med-Gemini-L 1.0. For brevity, we only display one representative search result in the example below.

\begin{tcolorbox} \footnotesize \textbf{Input} \\
\textit{Instruction} \\You are a medical expert answering a multiple choice question about medical knowledge.
To help you answer the question, you are given access to search results. \\

\textit{Question} \\A 20-year-old woman presents with menorrhagia for the past several years. She says that her menses ``have always been heavy'', and she has experienced easy bruising for as long as she can remember. Family history is significant for her mother, who had similar problems with bruising easily. The patient's vital signs include: heart rate 98/min, respiratory rate 14/min, temperature 36.1°C (96.9°F), and blood pressure 110/87 mm Hg. Physical examination is unremarkable. Laboratory tests show the following: platelet count 200,000/mm3, PT 12 seconds, and PTT 43 seconds. Which of the following is the most likely cause of this patient’s symptoms? (A) Hemophilia A (B) Lupus anticoagulant (C) Protein C deficiency (D) Von Willebrand disease
\\

\textit{Search Results} \\
Partial thromboplastin time (PTT) and activated partial thromboplastin time (aPTT) are used to test for the same functions; however, in aPTT, an activator is added that speeds up the clotting time and results in a narrower reference range. The aPTT is considered a more sensitive version of the PTT and is used to monitor the patient's response to heparin therapy. The reference range of the aPTT is 30-40 seconds. The reference range of the PTT is 60-70 seconds. In patients receiving anticoagulant therapy, the reference range is 1.5-2.5 times the control value in seconds. Critical values that should prompt a clinical alert are as follows: aPTT: More than 70 seconds (signifies spontaneous bleeding). PTT: More than 100 seconds (signifies spontaneous bleeding). eMedicine Logo. Next: Interpretation. A prolonged aPTT result may indicate the following : Congenital deficiencies of intrinsic system clotting factors such as factors VIII, IX, XI, and XII, including hemophilia A and hemophilia B (Christmas disease), two inherited bleeding disorders resulting from a deficiency in factors VIII and IX, respectively. Congenital deficiency of Fitzgerald factor (prekallikrein). Von Willebrand disease, which is the most common inherited bleeding disorder, affecting platelet function owing to decreased von Willebrand factor activity. Hypofibrinogenemia · Liver cirrhosis (the liver makes most of the clotting factors, including those that 
are vitamin K-dependent ones); diseases of the liver may result in an inadequate quantity of clotting
factors, prolonging the aPTT. Vitamin K deficiency: The synthesis of some clotting factors requires vitamin K, so vitamin K deficiency results in an inadequate quantity of intrinsic system and common pathways clotting factors, as a result the aPTT is prolonged. Disseminated intravascular coagulation (DIC): The clotting factors involved in the intrinsic pathway are consumed, prolonging the aPTT. \\
Source: \color{blue}https://emedicine.medscape.com/article/2085837-overview\color{black}

\end{tcolorbox}

\begin{tcolorbox} \footnotesize \textbf{Target} \\
The patient's symptoms of menorrhagia and easy bruising, along with a prolonged PTT, suggest a bleeding disorder.
Hemophilia A (option A) is an X-linked recessive disorder that typically presents with a prolonged PTT and normal PT. However, it is unlikely in this case as the patient is female and has a family history of her mother having similar symptoms.
Lupus anticoagulant (option B) can cause a prolonged PTT, but it is typically associated with thrombosis rather than bleeding.
Protein C deficiency (option C) can cause thrombosis, but it does not affect PTT.
According to the search results, ``Von Willebrand disease (vWD) is the most common inherited bleeding disorder, affecting platelet function owing to decreased von Willebrand factor activity.'' It can present with menorrhagia, easy bruising, and a prolonged PTT. The patient's family history of her mother having similar symptoms also supports this diagnosis.
Therefore, the most likely cause of this patient's symptoms is (D) Von Willebrand disease. \\
Answer: (D) Von Willebrand disease
\end{tcolorbox}

\paragraph{Uncertainty-guided search at inference} We design a novel, uncertainty-guided and iterative search process to improve Med-Gemini-L 1.0's generations at inference time. As displayed in the right panel of Figure \ref{fig:search_diagram}, each iteration consists of four steps: multiple reasoning path generation, uncertainty-based search invocation, uncertainty-guided search query generation, and finally search retrieval for prompt augmentation.
Note that while uncertainty-guided search at inference could potentially benefit multimodal settings, we only apply this approach to text-only benchmarks and leave multimodal exploration for future work.

\begin{enumerate}
    \item \textbf{Multiple reasoning path generation}: Given an input context prompt with a medical question, we generate multiple reasoning paths from Med-Gemini-L 1.0. For the first iteration, the prompt only consists of the instruction and question. For subsequent iterations, the prompt also includes search results from step (4) below. 
    \item \textbf{Uncertainty-based search invocation}: Given the multiple reasoning paths from step (1), we define an uncertainty measure based on the Shannon entropy of the answer choice distribution. Specifically, we calculate the probability of each answer choice by dividing its occurrence by the total number of responses, and apply the entropy based on the answer choice probabilities~\citep{horvitz1984diagnostic}. High entropy (model responses are more uniform across the different answer choices) indicates a high epistemic uncertainty. If the uncertainty for a question is higher than a defined threshold, we perform the uncertainty-guided search process in steps (3) and (4); otherwise, the majority vote answer is returned as the final answer.
    \item \textbf{Uncertainty-guided search query generation}: Given conflicting responses from step (1), we prompt Med-Gemini-L 1.0 to generate three search queries whose results would help resolve the conflict. Our motivation of conditioning on previously generated but conflicting responses is to retrieve search results that are directly targeted at resolving the model's uncertainty to the question.
    \item \textbf{Search retrieval}: The generated queries are then submitted to a web search engine, and the retrieved results are incorporated into Med-Gemini-L 1.0's input prompt for the next iteration, starting back at step (1). Augmenting the prompt with search results enables the model to refine its response by considering external relevant insights obtained from web search.
\end{enumerate}

\subsection{Multimodal understanding via fine-tuning and customized encoders}
To specialize Gemini's multimodal reasoning and conversational capabilities to the medical domain, we perform instruction fine-tuning of Gemini over a collection of domain-specific multimodal tasks following a similar procedure in prior works by~\cite{tu2024towardsmpm},~\cite{yu2022coca}, and~\cite{alayrac2022flamingo}. We use eight multimodal tasks across six datasets as shown in Table~\ref{tab:mm_ft}. A detailed description of the datasets is provided in the Appendix~\ref{app:mm_ft}.

\paragraph{Image-to-text multimodal fine-tuning} We use four image-to-text datasets from MultiMedBench \citep{tu2024towardsmpm, tanno2024consensus} including Slake-VQA~\citep{liu2021slake}, Path-VQA~\citep{he2020pathvqa}, MIMIC-CXR~\citep{johnson2019mimic,johnson2019mimicjpg}, PAD-UFES-20~\citep{pacheco2020pad}, in addition to the Radiology Objects in COntext (ROCO) dataset~\citep{pelka2018radiology}. Slake-VQA and Path-VQA include both open-ended and close-ended visual question answering tasks in radiology and pathology, respectively. ROCO contains radiology image captioning tasks spanning multiple imaging modalities including computed tomography (CT), ultrasound, X-ray [chest X-ray (CXR), fluoroscopy, mammography, angiography], positron emission tomography (PET) and magnetic resonance imaging (MRI). PAD-UFES-20 is a domain specific dataset with diagnostic labels and patient clinical information designed for dermatology image classification. MIMIC-CXR is a radiology dataset comprised of CXRs, their corresponding text reports, and a set of discrete labels that denote the presence of 13 abnormal radiological conditions derived using the CheXpert labeler~\citep{irvin2019chexpert} (e.g., pneumonia). We use this dataset to formulate CXR report generation and image classification tasks for fine-tuning.
For each task, we fine-tune Gemini 1.5 Pro by providing task-specific instructions as shown in~\Cref{fig:mm_ex}. The mixture ratio for each task is approximately proportional to the number of training samples in each dataset. The resulting model is Med-Gemini-M 1.5.

\paragraph{Augmenting health signal modalities with new modality encoders} We anticipate that integrating various health-related signals will significantly enhance medical models and treatment decisions. These signals include data from consumer wearables (e.g., long-term heart rate measurements, activity levels), genomic information, nutritional data (e.g., images of meals), and environmental factors (e.g., air quality measurements). As a proof-of-concept, we expand Med-Gemini's capability to process raw biomedical signals. Specifically, we develop Med-Gemini-S 1.0 by augmenting Gemini 1.0 Nano with a specialized encoder using a cross-attention mechanism based on Flamingo~\citep{alayrac2022flamingo} to answer questions directly taking a raw 12-channel electrocardiogram (ECG) waveform as input. We use a subset of labeled ECG examples from the ECG-QA dataset~\citep{oh2023ecgqa} and formulate the task as close-ended question answering with the instruction shown in Figure~\ref{fig:mm_ex}.

\subsection{Long-context processing via instruction prompting and chain-of-reasoning}
Many applications in medicine require the analysis of a large amount of information and the expertise to identify subtle details of the domain. As introduced before, Gemini models have breakthrough long-context capabilities. We assess medically-relevant long-context performance for Med-Gemini-M 1.5 by meaningfully processing large amounts of fine-grained information for two different medical applications: a ``needle-in-a-haystack'' retrieval task from lengthy EHR notes and records; and tasks requiring understanding of medical videos. We describe various prompting strategies and chain-of-reasoning to enable accurate recall and reasoning of information.

\paragraph{Chain-of-reasoning for long EHR understanding}
Searching and retrieving clinically-relevant information from long EHR notes and records is a common and important task in patient care but must be performed with high precision and recall to enhance clinician efficiency and reduce workload~\citep{jensen2012mining, ford2016extracting}. Clinicians frequently curate a summary of their patient's historical conditions, symptoms, or procedures (the ``problem list''), which can be time-consuming and challenging for individuals with lengthy medical records. Difficulty arises with multiple factors hindering effective information retrieval in EHRs. 

Firstly, classic query expansion and matching mechanisms encounter limitations due to textual similarities between conditions with similar taxonomies and the diverse information models used in EHRs (e.g. ``Miller'' vs. ``Miller Fisher syndrome'', ``Diabetic nephropathy'' vs. ``Diabetes mellitus''). Vocabulary inconsistency in and between EHR systems presents issues including variations in how medical terms are encoded, such as acronyms (``rx'' vs. ``prescription''),  misspellings, or synonyms for the same condition. Secondly, EHRs often contain heterogeneous data structure such as a checklist-style data template: ``[ ] cough [x] headache'', where a mention does not always indicate the presence of a medical condition. Thirdly, the context of a mention influences its interpretation. For example, the mention of the same condition in a patient's ``Family History'' compared to their ``Past Medical History'' could have different interpretations and implications for the patient's care. Lastly, polysemous acronyms in medical notes can lead to misinterpretations. 

These challenges motivate the need for AI systems to address the task of context-aware retrieval of subtle or rare conditions, medications, or procedure mentions from long EHR records - a practical benchmark for evaluating the utility of Med-Gemini in medicine. We setup the long-context EHR understanding task based on our prior work~\citep{feder2022building}, where we curate a set of long and challenging EHR cases from MIMIC-III~\citep{johnson2016mimic}, and formulate a subtle medical problem (condition/symptom/procedure) search-retrieval task over a collection of EHR notes and records, mimicking a clinically-relevant ``needle-in-a-haystack''~\citep{team2024gemini} problem. Details of the dataset and task curation procedure are described in~\Cref{app:lc_eval} and~\Cref{sec:lc-eval}.

To assess the long-context retrieval and reasoning capability of Med-Gemini-M 1.5, we aggregate the EHR notes across multiple visits from a single patient in each example and utilize the long-context window of the model with a two-step chain-of-reasoning approach (using only in-context learning). In the first step, we prompt Med-Gemini-M 1.5 to retrieve all mentions (snippets of evidence) related to the given problem (condition/symptom/procedure) with a one-shot demonstration. In the second step, we further prompt Med-Gemini-M 1.5 to determine the presence of the given problem entities based on the mentions retrieved. Details of the instruction prompts are shown in~\Cref{fig:ehr_ex} and~\Cref{sec:lc-eval}.

We use our prior heuristic-based annotation-aggregation method~\citep{feder2022building} as a baseline method for comparison with Med-Gemini-M 1.5. This heuristic-based method requires an extensive effort of manual feature engineering to determine the existence of a problem (condition/symptom/procedure) from a set of medical records. It is an ontology-dependent multiple-step process, which includes an annotation step that labels the problem in each EHR note, a rule-based selection step that selects mentions of problem entities with high confidence, and another rule-based aggregation step that aggregates all selected problem mentions to reach a final conclusion. Note that the manually crafted aggregation rules can only provide a limited coverage of all possible conditions, and therefore it requires additional engineering effort to expand coverage to new conditions. 

To curate a ``needle-in-a-haystack'' evaluation benchmark, we select medical conditions from a collection of EHR records with only one evidence snippet found in the aggregation step. We note that a mention of a condition in the EHR does not always mean the patient has that condition. This task enables us to assess Med-Gemini-M 1.5's ability to identify rarely documented and subtle conditions, symptoms, and procedures and reason accurately and holistically regarding their existence.

\paragraph{Instruction prompting for medical video understanding}
The understanding of surgical and procedural videos is a highly active research topic in medical AI. The advancing frontier of computer vision in semantic segmentation, object detection and tracking, and action classification has enabled new clinical applications such as surgical phase recognition, tool detection and tracking, and even surgical skill assessment~\citep{goodman2024analyzing}. 

Limited model context windows have hindered the ability for vision-language models to capture long-range dependencies and complex relationships within videos. Gemini's long-context capability offers a potential breakthrough for medical video understanding. By processing a whole video input, Med-Gemini-M 1.5 is able to identify visual patterns and understand actions and relationships between events across extended time frames.

To enable Med-Gemini-M 1.5 to understand medical videos, we employ zero-shot prompting with task-specific instructions as shown in~\Cref{fig:medvidqa_ex},~\Cref{fig:cholec80_cvs_exp}, and~\Cref{fig:avos_ex}. The goal is to enable the model to analyze the language query and video content, and perform the given task related to the input medical video---either localizing the relevant visual segment matching the query for the medical visual answer localization (MVAL) task~\citep{gupta2023dataset}, or identifying the surgical view in the video frames for the Critical View of Safety (CVS) assessment task~\citep{strasberg2010rationale,rios2023cholec80}.
More details on the medical video datasets and evaluation metrics are described in~\Cref{app:lc_eval} and~\Cref{sec:lc-eval}. 

%% file: evaluation.tex
\section{Evaluation}
We present evaluation benchmarks spanning (1) text-based reasoning, (2) multimodal, and (3) long-context processing tasks, demonstrating Med-Gemini's performance across a wide range of capabilities in medicine.

\subsection{Evaluation of advanced reasoning on text-based tasks}
We evaluate the medical reasoning capability of Med-Gemini-L 1.0 on three text benchmarks assessing clinical reasoning and the ability to retrieve information using web search to reduce uncertainty:
\begin{itemize}
    \item\textbf{MedQA (USMLE)}: a close-ended multiple-choice (4 options) dataset with 1273 USMLE style test questions curated by~\cite{jin2021disease}.
    \item \textbf{NEJM clinico-pathological conferences (NEJM CPC)}: a dataset comprising complex diagnostic case challenges in the medical journal, New England Journal of Medicine (NEJM) curated by~\cite{mcduff2023towards}.
    \item \textbf{GeneTuring}: a dataset that includes 600 open/close-ended QA pairs to evaluate genomic knowledge of LLMs \citep{hou2023geneturing}.
\end{itemize}

For MedQA, we follow the input-output format, and the evaluation method as described in~\cite{singhal2023large} using prediction accuracy as the metric. 
At inference, we go through four iterations of uncertainty-guided search.
Additionally, we ask board-certified primary care physicians (PCPs) from the US to relabel the MedQA test set. This enables us to identify questions with missing information such as plots or figures, labeling errors, and other potentially ambiguous questions with multiple possible correct answers~\citep{stutz2023evaluating}. Overall, this allows us to better characterize our performance on MedQA (USMLE). More details on this rating task can be found in Appendix \ref{app:medqa-relabeling}.

NEJM CPC evaluation is an open-ended diagnosis task. The input is a text-based, challenging clinico-pathological case (CPC) report, and the output is a differential diagnosis list, comprising 10 potential diagnoses. We use the top-1 and top-10 accuracy of identifying the correct diagnosis of the given challenging case, and use the same prompting procedures following~\cite{mcduff2023towards}.
At inference, we go through one iteration of uncertainty-guided search.

GeneTuring consists of 12 modules, each containing 50 open or close-ended QA pairs. We use the prediction accuracy as the evaluation metric, where the evaluation method and scoring technique for each module follow the methods described in \cite{hou2023geneturing}. 
In particular, we exclude from numerical evaluation, cases where the model outputs either do not directly answer or acknowledge limitations (i.e., abstained). At inference, we again go through only one iteration of uncertainty-guided search similar to NEJM CPC evaluation.

Beyond these benchmarks, we further evaluate Med-Gemini-M 1.0 on three challenging use cases that require long-form text generation. To this end, we conduct an expert evaluation where a panel of clinicians compare the responses of our model to those of other human experts via a side-by-side blinded preference comparison (more details are provided in Appendix~\ref{app:long_form_sxs}):

\begin{itemize}
    \item \textbf{Medical summarization}: Generate an after-visit summary (AVS) given de-identified history and physical (H\&P) notes. An AVS is a structured report that patients receive at the end of a medical appointment to summarize and guide their care journeys. 
    \item \textbf{Referral letter generation}: Generate a referral letter to another healthcare provider given a de-identified outpatient medical note that contains a recommendation for a referral.
    \item \textbf{Medical simplification}: Generate a plain language summary (PLS) given a technical abstract from a medical systematic review. A PLS should be written in plain English which can be understood by most readers without a university education \citep{cochrane-pls}.
\end{itemize}

\subsection{Evaluation of multimodal capabilities}
We evaluate Med-Gemini on seven multimodal visual question answering (VQA) benchmarks. For in-distribution evaluation, we choose four medical specialty datasets used in the instruction fine-tuning of Med-Gemini: PAD-UFES-20 (dermatology), Slake-VQA (radiology in English and Chinese) and Path-VQA (pathology) for Med-Gemini M 1.5, and ECG-QA (cardiology) for Med-Gemini S 1.0.

We also include three cross-specialty benchmarks for measuring out-of-box performance of Med-Gemini: NEJM Image challenge, USMLE-MM (multimodal), and MMMU-HM (health and medicine) datasets. These datasets are not used in any training or fine-tuning process. For this, we focus our evaluation on the Med-Gemini-L 1.0 model without any multimodal finetuning. 

Its worth noting that PAD-UFES-20, NEJM Image Challenge, USMLE-MM datasets, and most questions in MMMU-HM are close-ended VQA, i.e., multiple-choice question in a VQA setup. An overview of the selected datasets is presented in~\Cref{tab:mm_eval} and more details are in Appendix~\ref{app:mm_ft} and~\ref{app:mm_eval}.

We report prediction accuracy for all the close-ended multiple-choice VQA tasks, including NEJM Image Challenge, USMLE-MM, and PAD-UFES-20 6-class skin condition classification. We also follow the evaluation setup in~\cite{yue2023mmmu} to report accuracy for MMMU-HM. We use the exact-match accuracy for ECG-QA following~\cite{oh2023ecgqa}. For the open-ended VQA tasks (Slake-VQA and Path-VQA), we use the token-level F1 score following~\cite{tu2024towardsmpm}. 

We further showcase Med-Gemini-M 1.5's multimodal capability in multimodal medical diagnostic dialogue in two specialities - dermatology and radiology~\citep{tu2024towardsamie} - with qualitative evaluation  of the example dialogues by attending expert clinicians in these specialties. We note that these demonstrations indicate the "art of the possible", but that extensive further research and validation would be required before the consideration of deployment for a safety-critical use-case such as diagnostic assistance to a clinician.

\vspace{-5pt} 
\subsection{Evaluation of long-context capabilities on video and EHR tasks}
\label{sec:lc-eval}
We consider three tasks to demonstrate Med-Gemini-M 1.5's ability to seamlessly understand and reason over long context medical information (Table~\ref{tab:lc_eval}, details in Appendix~\ref{app:lc_eval}):
\begin{itemize}
    \item Long unstructured EHR notes understanding
    \item Medical instructional video QA
    \item Critical view of safety (CVS) assessment of surgical video
\end{itemize}

\paragraph{Long EHR understanding}
For the long-context EHR understanding task, we curate a MIMIC-III-Needle-in-a-Haystack task where the goal is to retrieve the relevant text spans of any mention of a given medical problem (condition/symptom/procedure) over a large collection of clinical notes in EHR and determine the existence of the condition by reasoning across the retrieved evidence.
Specifically, we curate 200 examples where each example consists of a collection of de-identified EHR notes selected from 44 unique ICU patients with a long medical history based on the following criteria:
\vspace{-8pt} 
\begin{itemize}
    \item Patients with long records: more than 100 medical notes (excluding structured EHR data). The length of each example ranges from 200,000 to 700,000 words. 
    \item In each example, the condition is mentioned only once across the collection of all EHR notes.
    \item Each sample has a single condition of interest.
\end{itemize}
\vspace{-8pt} 
The ground-truth label of each sample is a binary variable indicating whether a given problem entity of interest is present or not, obtained from the majority vote of three physician raters. Across the 200 test examples, the number of positive cases and negative cases are 121 and 79, respectively.

We compare Med-Gemini-M 1.5's one-shot in-context learning performance against the heuristic-based annotation-aggregation baseline method~\citep{feder2022building} in terms of precision and recall.
\vspace{-17pt} 
\paragraph{Video understanding}
We quantitatively evaluate Med-Gemini-M 1.5's long-context performance in the setting of video question-answering using three medical video tasks: two medical visual answer localization (MVAL) tasks using the Medical Instructional Video QA (MedVidQA) dataset~\citep{gupta2023dataset}, and the critical view of safety (CVS) assessment task on the Cholec80-CVS dataset~\citep{twinanda2016endonet,rios2023cholec80}.

The goal of MVAL is to identify specific video segments based on natural language descriptions (queries) given a video input. For MVAL, we benchmark the test set of MedVidQA for two video span prediction tasks, one using both the video input and subtitle text and the other one with only the video inputs. We follow~\cite{li2022towards,gupta2023dataset} using Intersection over Union (IoU) at the threshold of 0.3, 0.5, 0.7, and mean IoU (mIoU) as the evaluation metrics for the video span prediction tasks. IoU and mIoU are used to measure how much of the ground truth span overlaps with the predicted span.

We evaluate Med-Gemini-M 1.5's long-context capabilities in assessing the achievement of the Critical View of Safety (CVS) method in laparoscopic cholecystectomy (a keyhole operation to remove the gallbladder) videos. The CVS~\citep{strasberg2010rationale} is a recommended protocol used for secure identification of the cystic duct and cystic artery to minimize the risk of Bile Duct Injury (BDI), a significant injury associated with consequential postoperative morbidity and mortality, reduced long-term survival and impact on quality of life~\citep{way2003causes}. We evaluate the CVS assessment task on the public Cholec80 dataset~\citep{twinanda2016endonet} and Cholec80-CVS~\citep{rios2023cholec80} video clip annotations. Specifically, for each surgical video in the Cholec80 dataset, the Cholec80-CVS dataset provides annotations for video clips within the full video, where at least one CVS criteria is met. Each of those video clips is annotated with a score of 0, 1 or 2 for each of the three CVS criteria. All frames contained in a given video clip are considered to share the same annotation.
We evaluate the model's ability to predict which of the CVS criteria are met based on the whole video clip. We then compute the average accuracy of the answer against the Cholec80-CVS annotations across 572 annotated video clips. More details on the CVS task can be found in Appendix~\ref{app:lc_eval}.

Furthermore, to show the real-world capability of Med-Gemini-M 1.5 in capturing surgical actions in procedural videos, we qualitatively evaluate the surgical action recognition task using examples from the Annotated Videos of Open Surgery (AVOS) dataset~\citep{goodman2021real}, a video collection of open surgical procedures uploaded to the YouTube platform.

%% file: results.tex
\section{Results}
As introduced previously, we evaluate Med-Gemini's advanced reasoning, multimodal, and long-context capabilities across a wide range of medical benchmarks, both quantitatively and qualitatively. The array and diversity of tasks considered in this work is to the best of our knowledge, the most comprehensive for medical LLMs.
Further, our evaluations of Med-Gemini go beyond benchmarking of model capabilities and extend to tasks reflecting the potential for real-world utility, such as medical summarization, multimodal conversations, and surgical video understanding.

\subsection{Med-Gemini demonstrates advanced reasoning on text-based tasks}

\begin{table}[ht!]
\footnotesize
\centering
\resizebox{1.0\textwidth}{!}{
\begin{tabular}{cccccccc}
\toprule
\textbf{Task} & \textbf{Dataset} & \textbf{OOD} & \textbf{Metric} & \textbf{Med-Gemini-L 1.0} & \textbf{SoTA} & \textbf{SoTA method} & \textbf{Reference} \\
\midrule
Close-ended QA & MedQA &  & Accuracy & \textbf{91.1} & 90.2 & GPT-4 with MedPrompt & \cite{nori2023can} \\
Open-ended QA & NEJM CPC & \cmark & Top-1 accuracy & \textbf{30.7} & 29.2 & AMIE & \cite{mcduff2023towards} \\
 &  &  & Top-10 accuracy & \textbf{72.3} & 59.1 & AMIE & \cite{mcduff2023towards} \\ \hline
Gene name extraction & GeneTuring & \cmark & Accuracy & \textbf{86.0} & 85.0 & GPT-4 & \cite{hou2023geneturing} \\
Gene alias & GeneTuring & \cmark & Accuracy & \textbf{72.7} & 66.0 & GPT-4 & \cite{hou2023geneturing} \\
Gene name conversion & GeneTuring & \cmark & Accuracy & \textbf{100.0} & 85.0 & GPT-4 & \cite{hou2023geneturing} \\
Gene location & GeneTuring & \cmark & Accuracy & \textbf{83.0} & 61.0 & GPT-4 & \cite{hou2023geneturing} \\
SNP location & GeneTuring & \cmark & Accuracy & 0.0 & \textbf{5.00} & ChatGPT & \cite{hou2023geneturing} \\
Gene SNP association & GeneTuring & \cmark & Accuracy & 0.0 & 0.0 & GPT-4 & \cite{hou2023geneturing} \\
Protein-coding genes & GeneTuring & \cmark & Accuracy & \textbf{100.0} & 97.0 & GPT-4 & \cite{hou2023geneturing} \\
Gene disease association & GeneTuring & \cmark & Accuracy & 82.1 & \textbf{84.0} & GPT-4 & \cite{hou2023geneturing} \\
Gene ontology & GeneTuring & \cmark & Accuracy & \textbf{52.3} & 42.0 & GPT-4 & \cite{hou2023geneturing} \\
TF regulation & GeneTuring & \cmark & Accuracy & \textbf{65.3} & 62.0 & GPT-4 & \cite{hou2023geneturing} \\
Human genome DNA alignment & GeneTuring & \cmark & Accuracy & 0.0 & \textbf{7.0} & BioGPT & \cite{hou2023geneturing} \\
Multi-species DNA alignment & GeneTuring & \cmark & Accuracy & 12.5 & \textbf{20.0} & GPT-3 & \cite{hou2023geneturing} \\

\bottomrule
\end{tabular}
}
\caption{\footnotesize\textbf{Text-based evaluation.} Performance comparison of Med-Gemini-L 1.0 versus state-of-the-art (SoTA) methods. OOD: out-of-distribution dataset.}
\label{tab:text_perf}
\end{table}

As shown in~\Cref{tab:text_perf}, Med-Gemini-L 1.0 scores $91.1\%$ accuracy on MedQA (USMLE), a new SoTA, outperforming our previous Med-PaLM 2, by $4.5\%$, and the recent results augmenting GPT-4 with complex, specialized prompting - MedPrompt~\citep{nori2023can} by $0.9\%$. In contrast to MedPrompt, our principled approach leverages general web search in an uncertainty-guided framework that can be easily to extended to more complex scenarios beyond MedQA. 

\begin{figure}[ht!]
    \centering
    \begin{minipage}{\textwidth}
        \includegraphics[width=\textwidth]{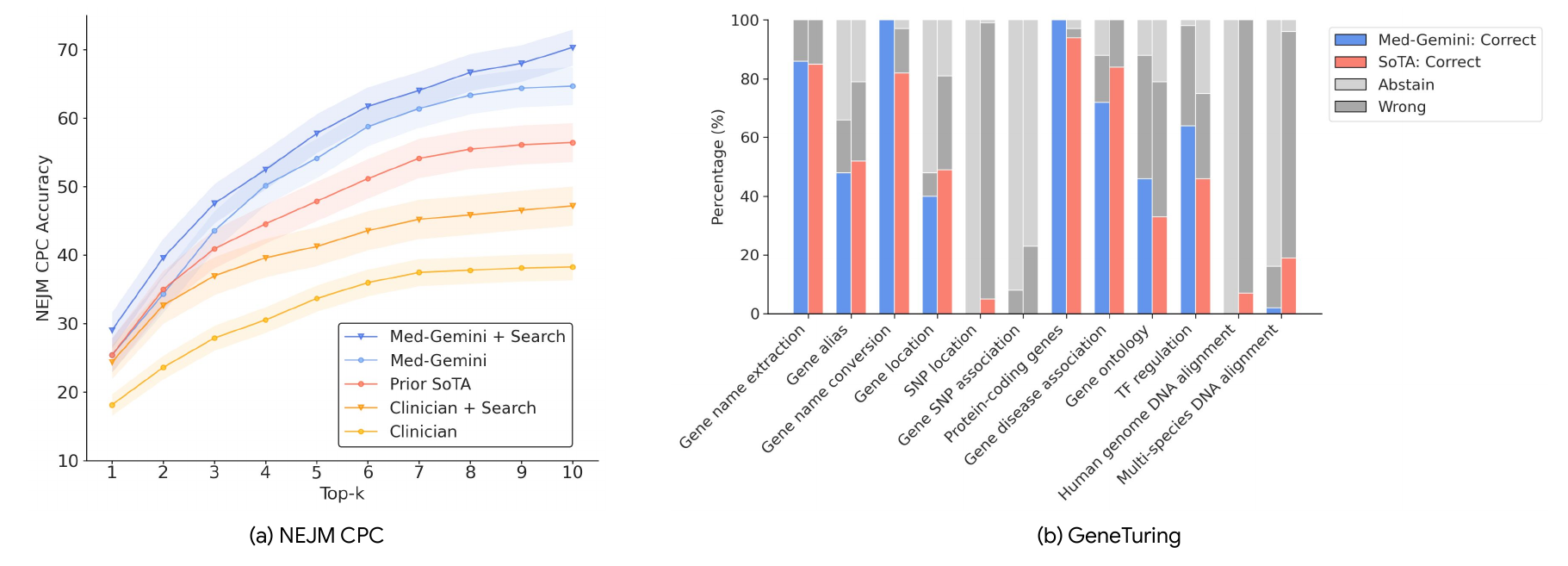}
    \end{minipage}
    \caption{\footnotesize\textbf{Generalization of Med-Gemini-L 1.0 with web search to two additional text-based benchmarks.} \textbf{(a)}: Comparison of Med-Gemini-L 1.0's top-k accuracy on the NEJM CPC benchmark with prior SoTA LLMs and clinicians, with and without search. \textbf{(b)}: Comparison between Med-Gemini-L 1.0 and SoTA models on the GeneTuring dataset modules. The bars represent the proportion of correct, incorrect, and abstention responses for each model.}
    \label{fig:text_results}
\end{figure}

As proof of generalization of our search integration, on the NEJM CPC complex diagnostic challenges benchmark, Med-Gemini-L 1.0 surpasses our previous SoTA AMIE model (which itself is better than GPT-4) \citep{mcduff2023towards} by $13.2\%$ on the top-10 accuracy as shown in~\Cref{fig:text_results}a. 

The same search strategy is also effective for genomics knoweledge tasks as shown in~\Cref{tab:text_perf}. Med-Gemini-L 1.0 outperforms the SoTA models reported in \cite{hou2023geneturing} on seven GeneTuring modules including \textit{Gene name extraction}, \textit{Gene alias}, \textit{Gene name conversion}, \textit{Gene location}, \textit{Protein-coding genes}, \textit{Gene ontology} and \textit{TF regulation}.
We also compare model abstention across the 12 modules in~\Cref{fig:text_results}b.
It is worth noting that GeneGPT~\citep{jin2024genegpt} achieves higher scores through specialized web APIs, while our comparison focuses on prior models from \cite{hou2023geneturing} that utilize general web search similar to our model.

\begin{figure}[ht!]
    \centering
    \includegraphics[width=\textwidth]{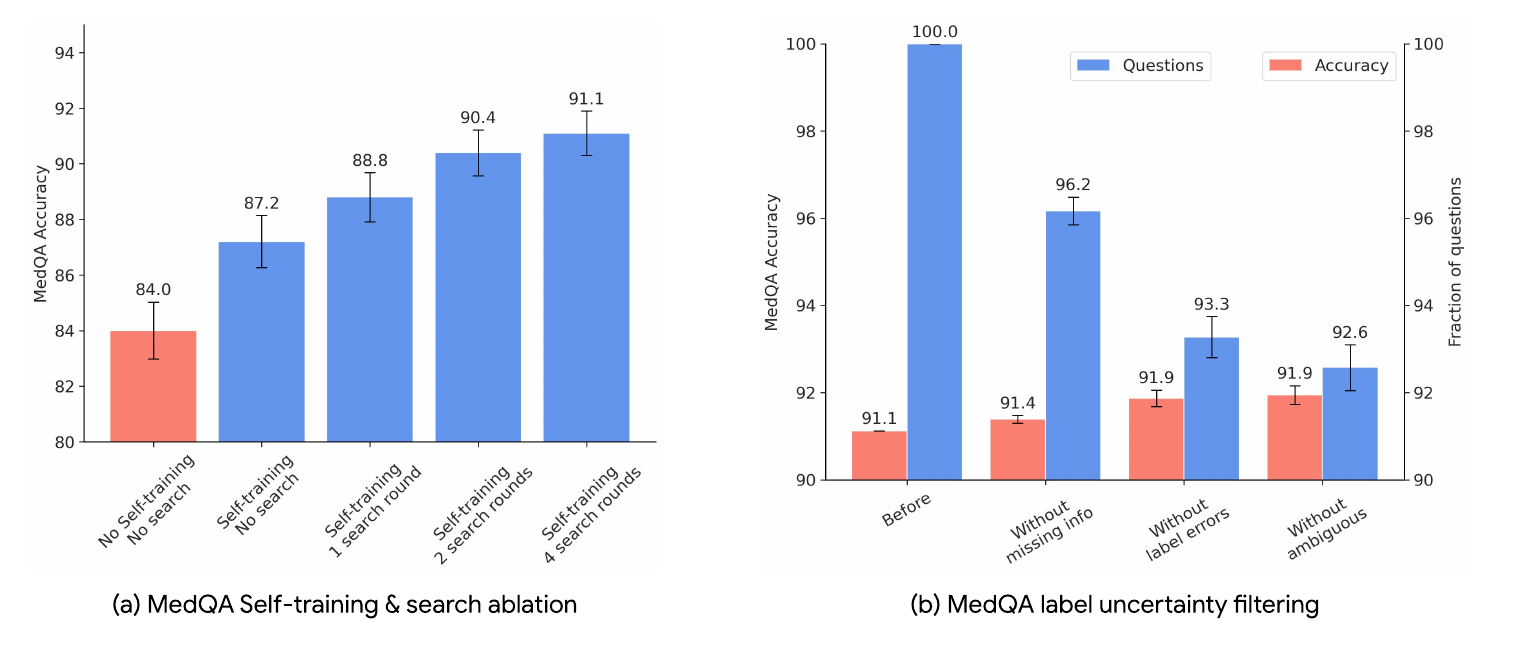}
    \caption{\footnotesize\textbf{Ablation analysis and label uncertainty on MedQA.} \textbf{(a)}: Impact of self-training and uncertainty-guided search on Med-Gemini-L 1.0's accuracy on MedQA. Self-training and each round of search contribute to significant performance improvements. \textbf{(b)}: Med-Gemini-L 1.0's accuracy (blue) and remaining questions (red) on MedQA after re-labeling by at least three US physicians per question. Filtering questions with missing information, label errors, or ambiguous groundtruth further improves accuracy. The error bars correspond to standard error across cases in (a) and standard deviation across bootstrapped annotations in (b).}
    \label{fig:medqa_ablations}
\end{figure}

\paragraph{Ablation analysis}
To understand the impact of self-training and uncertainty-guided search on performance, we compare Med-Gemini-L 1.0's performance with and without self-training, along with varying number of rounds of uncertainty-guided search for MedQA (USMLE). As shown in~\Cref{fig:medqa_ablations}a, Med-Gemini-L 1.0's performance improves considerably with self-training (a gain of $3.2\%$ in accuracy), and improves with each round of search from $87.2\%$ up to $91.1\%$. 
Similarly, for the NEJM CPC benchmark,~\Cref{fig:text_results}a shows a $4.0\%$ improvement for top-10 accuracy when we add search at inference. 
In~\Cref{app:cpc}, we additionally show performance on NEJM CPC stratified by four specialities. 

\paragraph{Revisiting MedQA (USMLE) labels}
MedQA (USMLE) is a popular benchmark for assessing the capabilities of LLMs in the medical domain. 
However, some MedQA test questions have missing information such as figures or lab results, and potentially outdated ground-truth answers. 
To address these concerns, we conduct a complete relabeling of the MedQA (USMLE) test set. Specifically, we recruit at least three US physicians to re-annotate each question, asking them to answer the question and evaluate the provided ground-truth answer. We also ask them to identify if there was any missing information in the questions. Following~\cite{stutz2023evaluating}, we characterize the questions to exclude due to missing information or label errors by bootstrapping votes from committees of three raters per question. We additionally identify \emph{ambiguous} questions as those allowing multiple correct answers (more details can be found in Appendix \ref{app:medqa-relabeling}).

~\Cref{fig:medqa_ablations}b shows that, on average across bootstrapped committees, $3.8\%$ of questions include missing information, following the unanimous vote of bootstrapped committees. Additionally, $2.9\%$ likely include label errors. Another $0.7\%$ are ambiguous. Excluding these questions is supported by high inter-rater agreement of $94\%$, $87.6\%$, and $94.6\%$, respectively. Importantly, Med-Gemini-L 1.0's mistakes can be attributed disproportionately to these questions; our entropy-based uncertainty score also tends to be higher on these question (t-test, $p$-value=0.033). Filtering both types improves accuracy from $91.1\%$ to $91.8\%$ $\pm$ $0.2\%$. Using majority instead of unanimous votes further improves accuracy to $92.9\%$ $\pm$ $0.38\%$ by discarding up to $20.9\%$ of the uncertain questions.

\subsubsection{Performance on long-form medical text generation}
Med-Gemini-M 1.0 demonstrates the ability to generate long-form text for three challenging real-world use cases - after-visit clinical summaries, doctor referral letter generation and medical simplification. In side-by-side comparisons, Med-Gemini-M 1.0's responses are considered as good or better than expert responses more than half the time by clinician raters across the three tasks (Figure~\ref{fig:long_form_sxs}). For more task details, see Appendix~\ref{app:long_form_sxs}. Notably for the referral letter generation task, the model generated letters are preferred or tied with experts across all the samples evaluated.

\begin{figure}[ht!]
    \centering
    \includegraphics[scale=0.35]{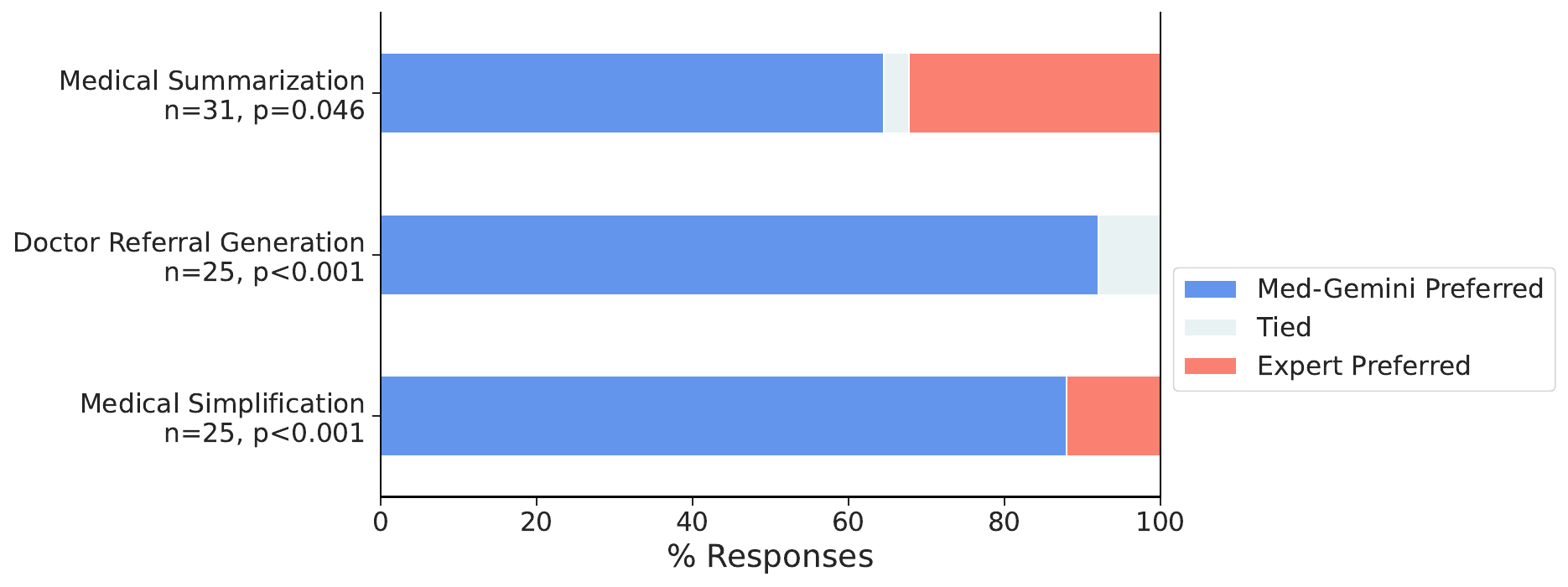}
    \caption{\footnotesize\textbf{Evaluation of Med-Gemini-M 1.0 on long-form text-based tasks via side-by-side comparison with experts.} The tasks considered include generation of after-visit summaries, referral letters and simplified summaries of medical systematic reviews. Evaluation was performed by clinician raters. P-values are used to denote whether the rate at which Med-Gemini-M 1.0 is preferred or tied with experts is $0.5$ (two-sided t-test).}
    \label{fig:long_form_sxs}
\end{figure}

\subsection{Med-Gemini demonstrates multimodal understanding across diverse tasks}

Our Med-Gemini models surpass, or perform competitively, with the state-of-the-art methods across seven medical multimodal benchmarks (See~\Cref{tab:mm_perf}). We provide representative input and output examples for the multimodal tasks in Figure~\ref{fig:mm_ex} for illustration.

In particular, Med-Gemini-L 1.0 reaches SoTA on three out-of-distribution close-ended VQA tasks---NEJM Image Challenge, multimodal USMLE sample questions (USMLE-MM), and the health \& medicine subset of MMMU (MMMU-HM), outperforming GPT-4V by $8.7\%$, $13.1\%$, and $2.6\%$, respectively. Meanwhile, Med-Gemini-M 1.5 outperforms our previous multimodal models, Med-PaLM M~\citep{tu2024towardsmpm} on Path-VQA by $2.0\%$ in token F1 score, and Med-Gemini-S 1.0 outperforms the previous SoTA for ECG-QA (GPT-4 with SE-WRN) by $6.1\%$ on macro-averaged accuracy across ECG question types~\citep{oh2023ecgqa}. Med-Gemini-M 1.5 also performs competitively on Slake-VQA and PAD-UFES-20 compared to the previous SoTA method (Med-PaLM M) but does not reach SoTA. 

\begin{table}[ht!]
\footnotesize
\centering
\resizebox{1.0\textwidth}{!}{
\begin{tabular}{cccccccc}
\toprule
\textbf{Task} & \textbf{Dataset} & \textbf{Multimodal fine-tuned} & \textbf{Metric} & \textbf{Med-Gemini} & \textbf{SoTA} & \textbf{SoTA method} & \textbf{Reference} \\
\midrule
Close-ended VQA & NEJM Image Challenge &  & Accuracy & \textbf{69.7$^*$} & 61.0 & GPT-4V & \cite{buckley2023accuracy} \\
Close-ended VQA & USMLE-MM &  & Accuracy & \textbf{93.5$^*$} & 80.4 & GPT-4V & Reproduced \\
Close/open-ended VQA & MMMU-HM &  & Accuracy & \textbf{67.3$^*$} & 64.7 & GPT-4V & \cite{yue2023mmmu} \\ 
Close-ended Signal QA& ECG-QA & \cmark & Accuracy & \textbf{57.7$^\ddagger$} & 51.6 & GPT-4 with SE-WRN & \cite{oh2023ecgqa} \\
Open/Close-ended VQA & Slake-VQA & \cmark & Token F1 & 87.5$^\dagger$ & \textbf{89.3} & Med-PaLM M & \cite{tu2024towardsmpm} \\
Open/Close-ended VQA & Path-VQA & \cmark & Token F1 & \textbf{64.7$^\dagger$} & 62.7 & Med-PaLM M & \cite{tu2024towardsmpm} \\
Classification & PAD-UFES-20 6-class & \cmark & Accuracy & 85.9$^\dagger$ & \textbf{88.0} & Med-PaLM M & \cite{tu2024towardsmpm} \\
Classification & PAD-UFES-20 6-class & \cmark & Accuracy & 78.8$^\dagger$ & N/A & N/A & New Split \\
\bottomrule
\end{tabular}
}
\caption{\footnotesize\textbf{Multimodal evaluation.} Performance comparison of Med-Gemini versus state-of-the-art (SoTA) methods. $*$ denotes the performance of Med-Gemini-L 1.0, $\dagger$ denotes the performance of Med-Gemini-M 1.5, and $\ddagger$ denotes the performance of Med-Gemini-S 1.0.}
\label{tab:mm_perf}
\end{table}

Note that we have evaluated PAD-UFES-20 on two different data split setups. We first evaluate on the Med-PaLM M split (the image-level split) for a direct, fair comparison against the previous SoTA method. In addition, we also report our model's performance on a new split, which is a split at the patient level (\Cref{tab:mm_perf}).

For USMLE-MM, our model achieves accuracies of $89.5\%$, $92.9\%$, $100.0\%$ for USMLE step 1 questions (n=19), step 2 (n=14), and step 3 (n=13), respectively.

In aggregate across these seven benchmarks, Med-Gemini improve over GPT-4V by an average relative margin of 44.5\%. Note that for the USMLE-MM, PADS-UFES-20 and Slake-VQA datasets, we report reproduced GPT-4V results using public APIs and the same prompt used for the corresponding Med-Gemini model.

\subsubsection{Preview of multimodal dialogue capabilities}
To extend beyond multimodal benchmarks, we demonstrate the potential for future real-world utility of Med-Gemini through hypothetical multimodal medical dialogues across two specialities. 

\begin{figure}[ht!]
    \centering
    \includegraphics[height=0.755\textheight,keepaspectratio]{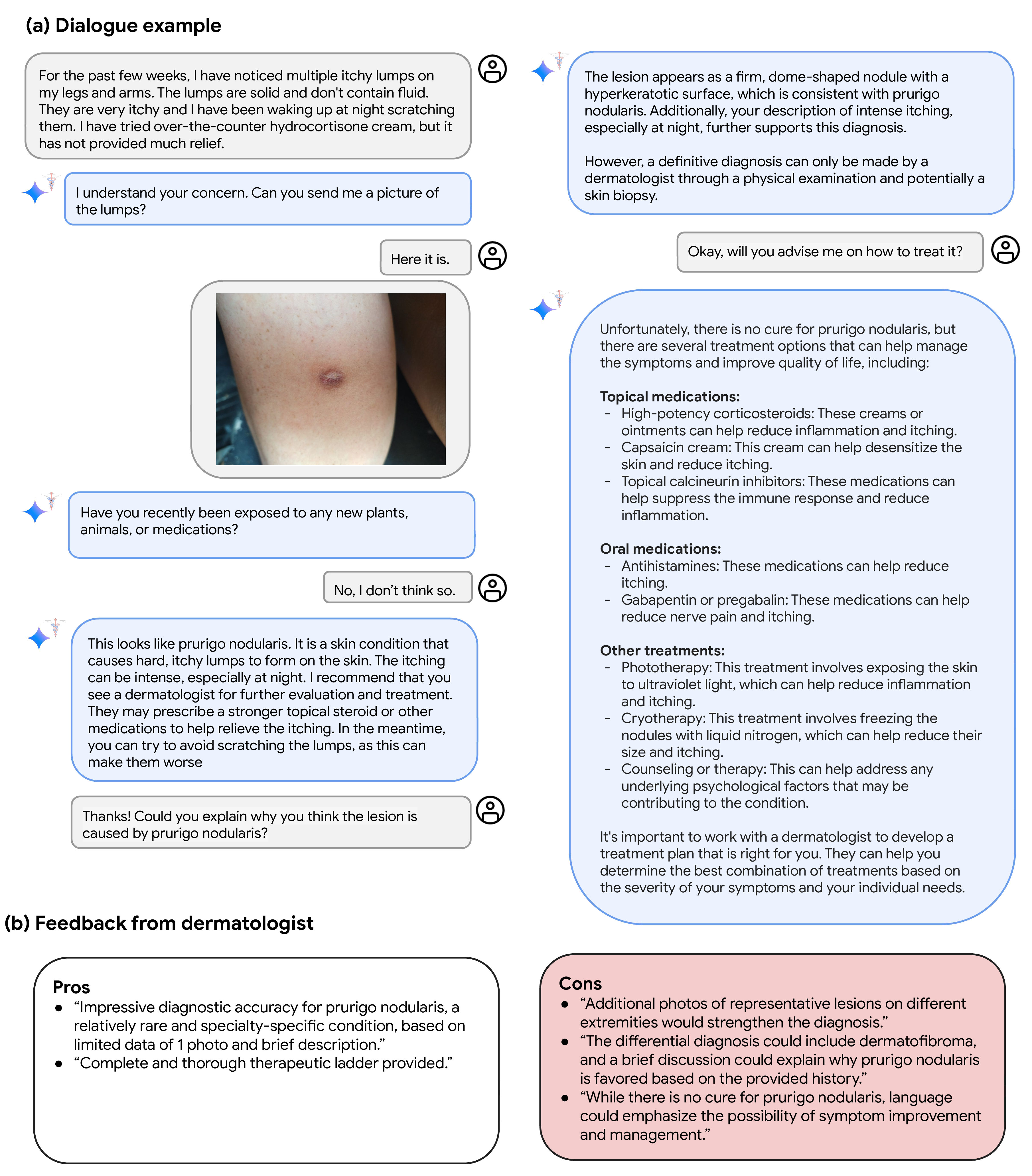}
    \caption{\footnotesize\textbf{Example of a hypothetical multimodal diagnostic dialogue with Med-Gemini-M 1.5 in a dermatology setting.} \textbf{(a)} The user interacts with the our multimodal model, Med-Gemini-M 1.5, acting as a patient based on a case from SCIN \citep{ward2024crowdsourcing}, an external dataset that is not included in the fine-tuning mixture. This system would not be appropriate for use in real-world diagnostic tasks without considerable further research and development. Nonetheless, this example shows indicatively the future initial promise of how the combination of the conversational capability inherited from the original Gemini model and the newly acquired multimodal medical knowledge from fine-tuning can potentially enable attributes of a rich multi-turn diagnostic dialogue. In this interaction, Med-Gemini-M 1.5 asks for a picture when it is not provided (\textit{multimodal information acquisition}), arrives at the right diagnosis efficiently (\textit{open-ended diagnosis}), explains the reasoning by integrating the relevant visual features and other gathered patient symptoms (\textit{explainability}), answers questions about treatment options while deferring aptly to experts for the final decision. \textbf{(b)} Shows feedback collected from dermatologists to evaluate qualitative the contents of the diagnostic dialogue, specifically asking them to comment on both positive and negative aspects.}
    \label{fig:dialogue-derm}
\end{figure}

\Cref{fig:dialogue-derm} illustrates an out-of-distribution setting where the dermatology image comes from a dataset~\citep{ward2024crowdsourcing} not used in the multimodal fine-tuning mixture. The user first asks Med-Gemini-M 1.5 about itchy lumps on their legs and arms; our model then asks the user to share an image of the lumps; after the user provides the image of their suspicious lesion, the model asks a follow-up question and continues to provide a correct diagnosis of prurigo nodularis, and recommends next steps and potential treatment options.

\begin{figure}[ht!]
    \centering
    \includegraphics[height=0.78\textheight,keepaspectratio]{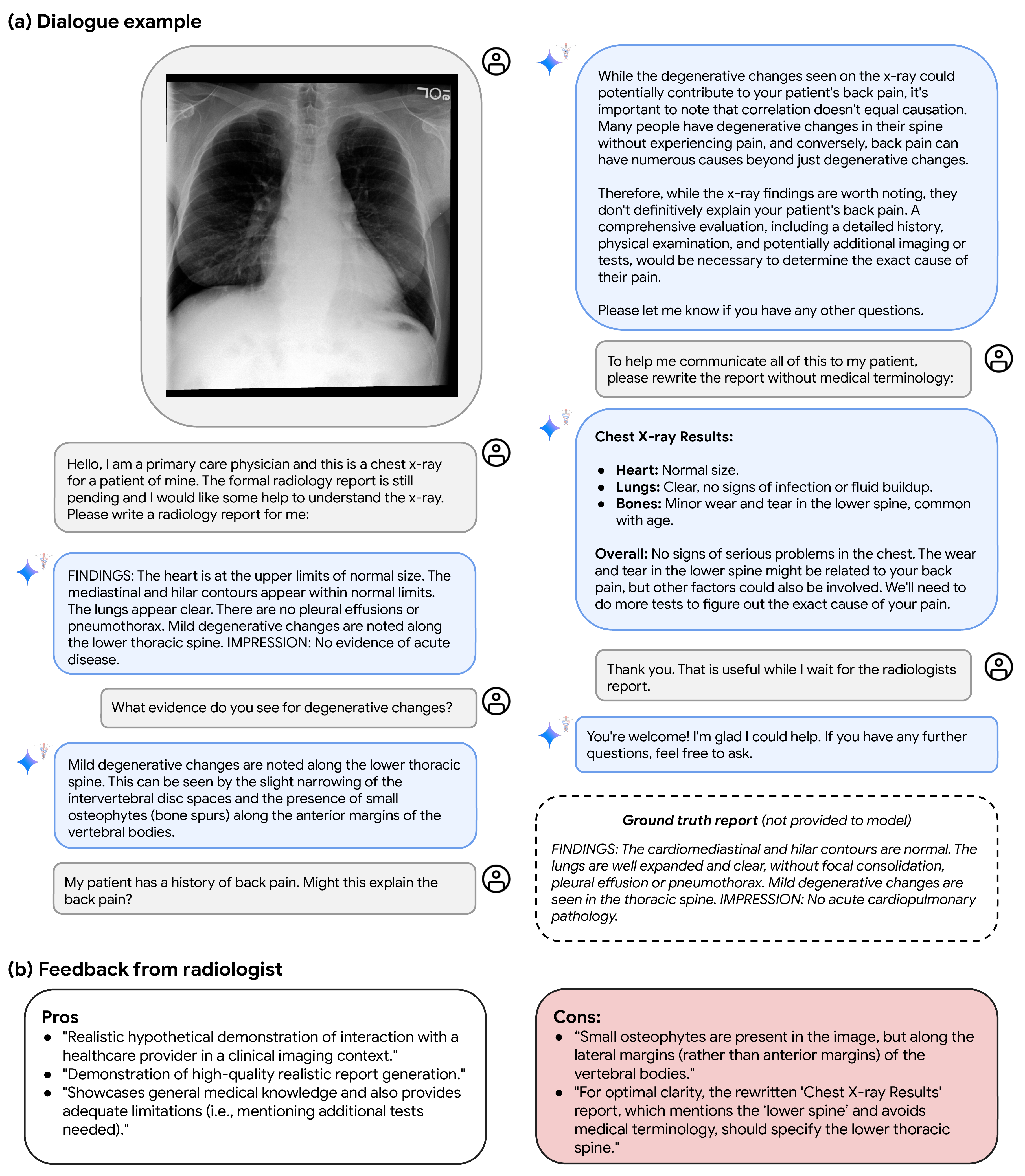}
    \caption{\footnotesize\textbf{Example of a hypothetical multimodal diagnostically-assistive dialogue with Med-Gemini-M 1.5 in the radiology setting.} 
    \textbf{(a)} In this interaction, Med-Gemini-M 1.5 demonstrates its ability to analyze a chest X-ray (CXR) and conduct a hypothetical realistic dialogue with a primary care physician. As above, Med-Gemini-M 1.5 is not suitable for this real-world use without further research. However, this example demonstrates initial promise, wherein Med-Gemini-M 1.5 identifies mild degenerative changes along the spine and can answer questions about the reasoning which led to this finding, demonstrate general medical knowledge about degenerative disk disease and distinguish between correlation and causation in relation to a patient history of back-pain. Finally, in this example Med-Gemini-M 1.5 is able to explain its findings in layperson's terms, demonstrating its potential for facilitating patient understanding and communication in clinical settings. The ground truth report for this CXR is provided. \textbf{(b)} Feedback from a radiologist about the quality of this radiology dialogue.} 
    \label{fig:dialogue-cxr}
\end{figure}

In \Cref{fig:dialogue-cxr}, we show a radiology dialogue example with the example image coming from the test set of the MIMIC-CXR dataset. Med-Gemini-M 1.5 demonstrates the ability to interact with a primary care provider to analyze a CXR, identify degenerative disk disease, discuss the differences between causation and correlation with a patient history of back-pain, suggest follow-up investigations to establish the cause of back pain and provide a report using non-technical language to facilitate patient understanding and communication. We observe some variability in Med-Gemini-M 1.5's responses depending on the prompt (for instance, for some prompts the report will not list mild degenerative changes especially if prompted to focus on some other anatomical feature). A full quantification of Med-Gemini-M 1.5's multimodal dialogue capability and variability is beyond the scope of this work, but nonetheless these qualitative examples illustrate Med-Gemini-M 1.5's ability to support conversations about medical knowledge grounded on multimodal sources, a potentially useful attribute for applications considering user-AI and clinician-AI interaction. Real-world exploration of these use-cases would require considerable further development and validation to build upon these signs of early promise.

\subsection{Med-Gemini shows long-context processing capability on long EHR and video tasks}
Finally, we evaluate the long-context capability of Med-Gemini-M 1.5 via the ``needle-in-a-haystack'' medical condition retrieval task from long EHRs as well as three medical video tasks (two MAVL and one CVS assessment of surgical videos).

\begin{figure}[htbp]
    \centering
    \includegraphics[width=\textwidth]{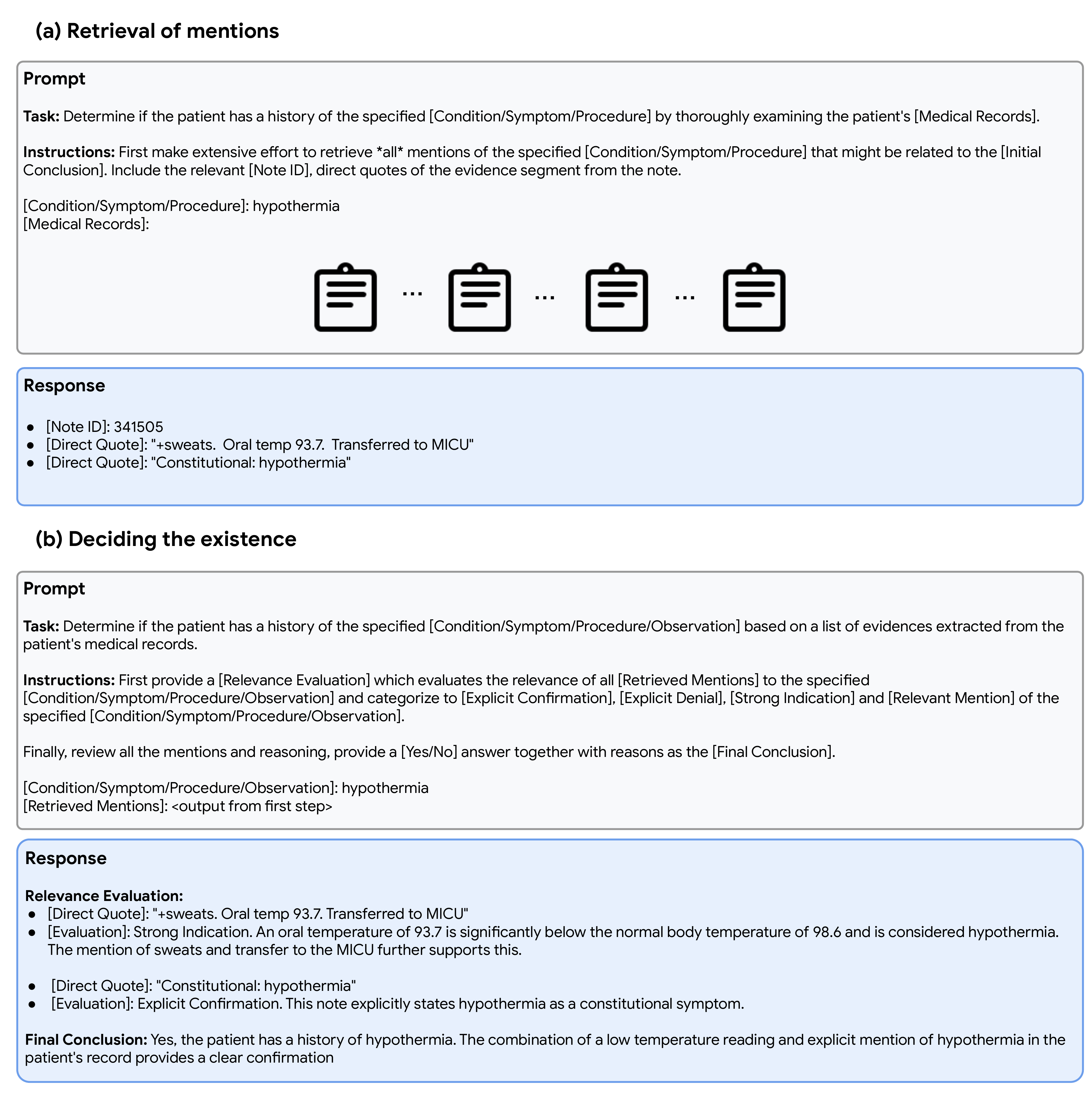}
    \caption{\footnotesize\textbf{Example of Med-Gemini-M 1.5's long-context capabilities on long EHR understanding (MIMIC-III Needle-in-a-Haystack).} Med-Gemini-M 1.5 performs a two-step process for determining whether a patient has a history of a specific condition based on their extensive EHR records. \textbf{(a)} Step 1 (Retrieval): Med-Gemini-M 1.5 identifies all mentions of ``hypothermia'' within the EHR notes, providing direct quotes [e.g., ``+sweats. Oral temp 93.7. Transferred to medical intensive care unit (MICU)''] and note IDs for each mention. \textbf{(b)} Step 2 (Deciding the existence): Med-Gemini-M 1.5 then evaluates the relevance of each retrieved mention, categorizing them as explicit confirmation, strong indication, or relevant mention of hypothermia. Based on this analysis, the model concludes that the patient does have a history of hypothermia, providing clear reasoning for its decision.
    }
    \label{fig:ehr_ex}
\end{figure}

We demonstrate the utility of Med-Gemini-M 1.5 on the correct identification of rare and subtle problem entity (condition/symptom/procedure) in long EHR notes. The average precision and recall between Med-Gemini-M 1.5 and the baseline method are shown in~\Cref{tab:lc_perf} (confidence intervals in~\Cref{tab:lc_ehr}). Encouragingly, we observe that Med-Gemini-M 1.5's one-shot ability is on-par with a carefully-tuned heuristic-based annotation-aggregation baseline approach, which is highly task-dependent. The in-context learning capability of Med-Gemini-M 1.5 to process long documents or records can easily generalize to novel problem settings without the need of extensive manual engineering. We provide an illustrative example of the prompt used, along with our model's response in Figure~\ref{fig:ehr_ex}.  We attempt to benchmark GPT-4 on this task but the average context token length in this dataset significantly exceeds the maximum context window supported in the public APIs.

\begin{table}[ht!]
\footnotesize
\centering
\resizebox{1.0\textwidth}{!}{
\begin{tabular}{cccccccc}
\toprule
\textbf{Task} & \textbf{Dataset} & \textbf{OOD} & \textbf{Metric} & \textbf{Med-Gemini} & \textbf{SoTA} & \textbf{SoTA method} & \textbf{Reference} \\
\midrule
EHR Needle-in-a-Haystack & MIMIC-III & \cmark & Precision & 0.77 & \textbf{0.85} & Annotation+Aggregation & \cite{feder2022building} \\
 &  &  & Recall & \textbf{0.76} & 0.73 & Annotation+Aggregation & \cite{feder2022building} \\
 &  &  & F1 & 0.77 & \textbf{0.78} & Annotation+Aggregation & \cite{feder2022building} \\
\midrule
Video QA (video-only) & MedVidQA & \cmark & IoU@0.3 & \textbf{60.8} & 32.9 & RaNet & \cite{li2022towards} \\
 &  &  & IoU@0.5 & \textbf{43.2} & 20.6 & RaNet & \cite{li2022towards} \\
 &  &  & IoU@0.7 & \textbf{31.0} & 15.5 & RaNet & \cite{li2022towards} \\
 &  &  & mIoU & \textbf{43.4} & 27.5 & RaNet & \cite{li2022towards} \\
Video QA (video+subtitle) & MedVidQA & \cmark & IoU@0.3 & \textbf{84.4} & 80.7 & MutualSL & \cite{weng2023visual} \\
 &  &  & IoU@0.5 & \textbf{72.9} & 61.9 & MutualSL, VPTSL & \cite{weng2023visual,li2022towards} \\
 &  &  & IoU@0.7 & \textbf{54.7} & 44.5 & VPTSL & \cite{li2022towards} \\
 &  &  & mIoU & \textbf{65.8} & 58.3 & MutualSL & \cite{weng2023visual} \\
CVS assessment & Cholec80-CVS & \cmark & Accuracy & 55.2 & \textbf{67.0} & ResNet3D & Reproduced \\
\bottomrule
\end{tabular}
}
\caption{\footnotesize\textbf{Long-context evaluation.} Performance comparison of Med-Gemini-M 1.5 versus the state-of-the-art (SoTA) methods. Note that 7 out of 155 questions in MedVidQA are not answerable due to YouTube video access (private, removed). We mark these tasks and benchmarks as out-of-distribution (OOD) because all the evaluation here is only with in-context learning and no fine-tuning}
\label{tab:lc_perf}
\end{table}

\begin{figure}[hbtp]
    \centering
    \includegraphics[width=\textwidth]{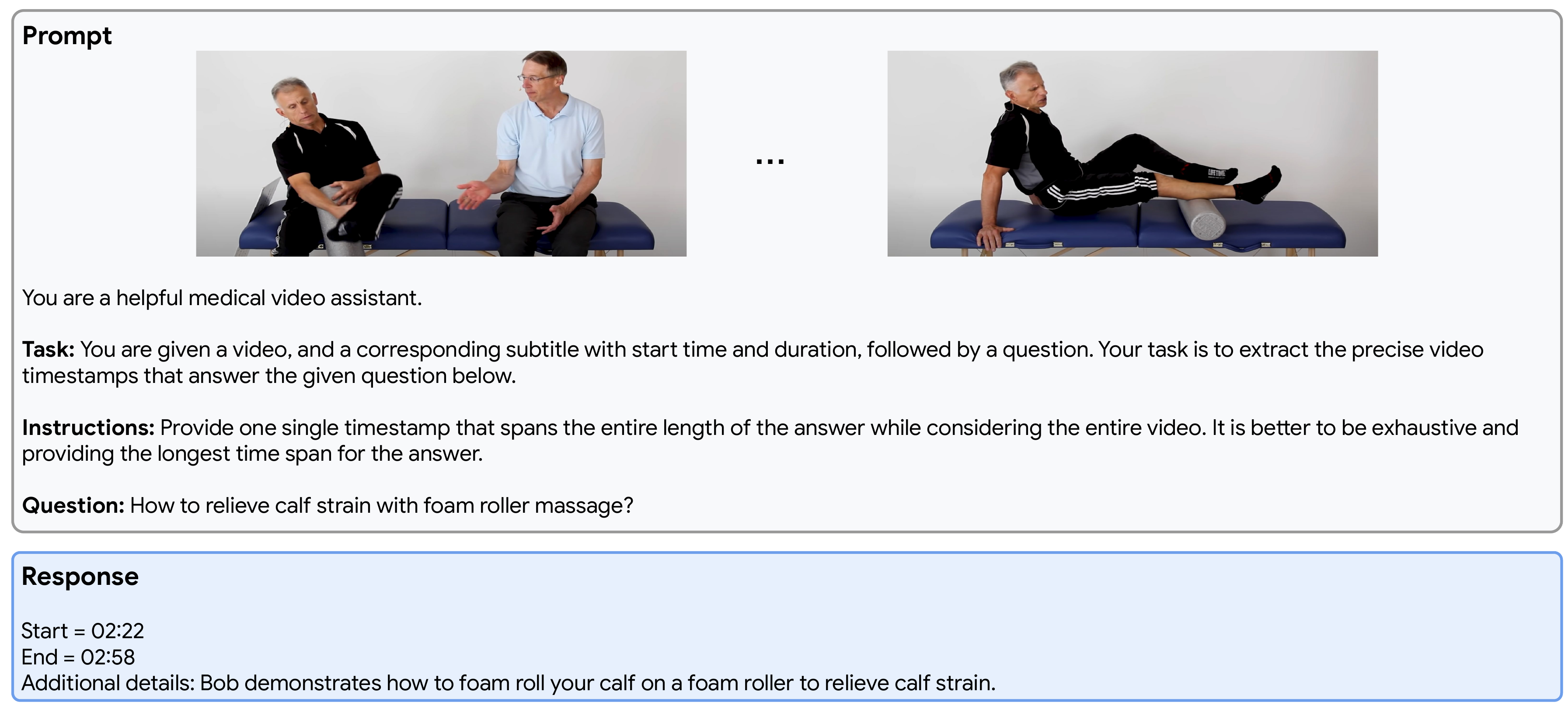}
    \vspace{-10pt}
    \caption{\footnotesize\textbf{Example of Med-Gemini-M 1.5's long-context capabilities on medical instructional videos.} Med-Gemini-M 1.5 analyzes a video from the Medical Video Question Answering (MedVidQA) dataset to answer a specific question about relieving calf strain. The model identifies the relevant video segment (02:22-02:58) where the physical therapist explains and demonstrates the exercise for this condition. The MedVidQA ground truth time span annotation is 02:22-03:00.
    }
    \label{fig:medvidqa_ex}
\end{figure}

Med-Gemini-M 1.5 also achieves SoTA performance on two MedVidQA MAVL tasks (one using both video and subtitles and the other being video only), outperforming the non-LLM based SoTA models which require considerable be-spoke tuning. We note that 7 questions in MedVidQA are not answerable due to YouTube video access (private, removed). Our results therefore are reported based on the remaining 148 questions. Details are shown in~\Cref{tab:lc_perf}.
We provide an illustrative example of the prompt used, along with our model's response in~\Cref{fig:medvidqa_ex}. While evaluating MedVidQA, we also observe that the embedded captions can significantly aid the model's understanding.
Future research could explore how to optimize the use of multimodal video data, including images, text, and audio, for further improvements in video understanding. We attempt to benchmark GPT-4V on these tasks but once again run into context length limitations for most of the videos using the public APIs.

\begin{figure}[hb]
    \centering
    \includegraphics[width=\textwidth]{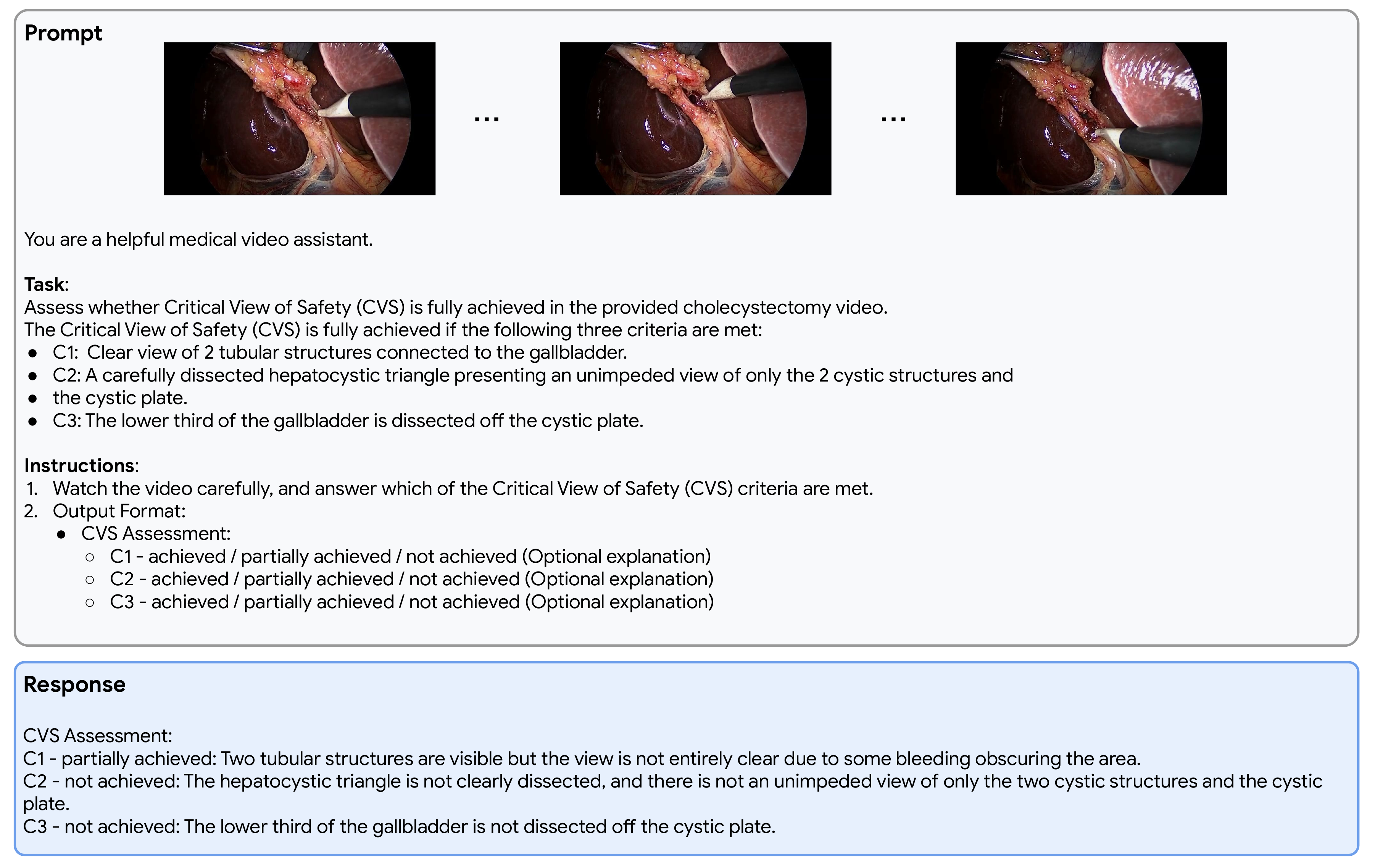}
    \caption{\footnotesize\textbf{Example of Med-Gemini-M 1.5's long-context capabilities on a surgical video.} Med-Gemini-M 1.5 analyzes a video clip from the Cholec80 dataset to assess achievement of the Critical View of Safety (CVS) during a laparoscopic cholecystectomy (a keyhole operation to remove the gallbladder). The model assesses whether the three criteria defining the achievement of the CVS are met, with a detailed explanation for each criterion separately.}
    \label{fig:cholec80_cvs_exp}
\end{figure}

For the CVS assessment of the laparoscopic cholecystectomy video task, Med-Gemini-M 1.5 outperforms GPT-4V by 21\%. However, we observe that the supervised baseline using a ResNet3D architecture performs better.
Further investigations on prompting strategies or instruction fine-tuning may be required to improve the task performance of our models.
We provide an illustrative example of the prompt used, along with our model's response in~\Cref{fig:cholec80_cvs_exp}.

\subsubsection{Applications of long-context capabilities in biomedicine}

In addition to quantitative results, we further preview the potentials of the long-context capabilities in medical education, facilitating clinician interaction with EHR systems and biomedical literature review and summarization.

\begin{figure}[hbtp]
    \centering
    \includegraphics[width=\textwidth]{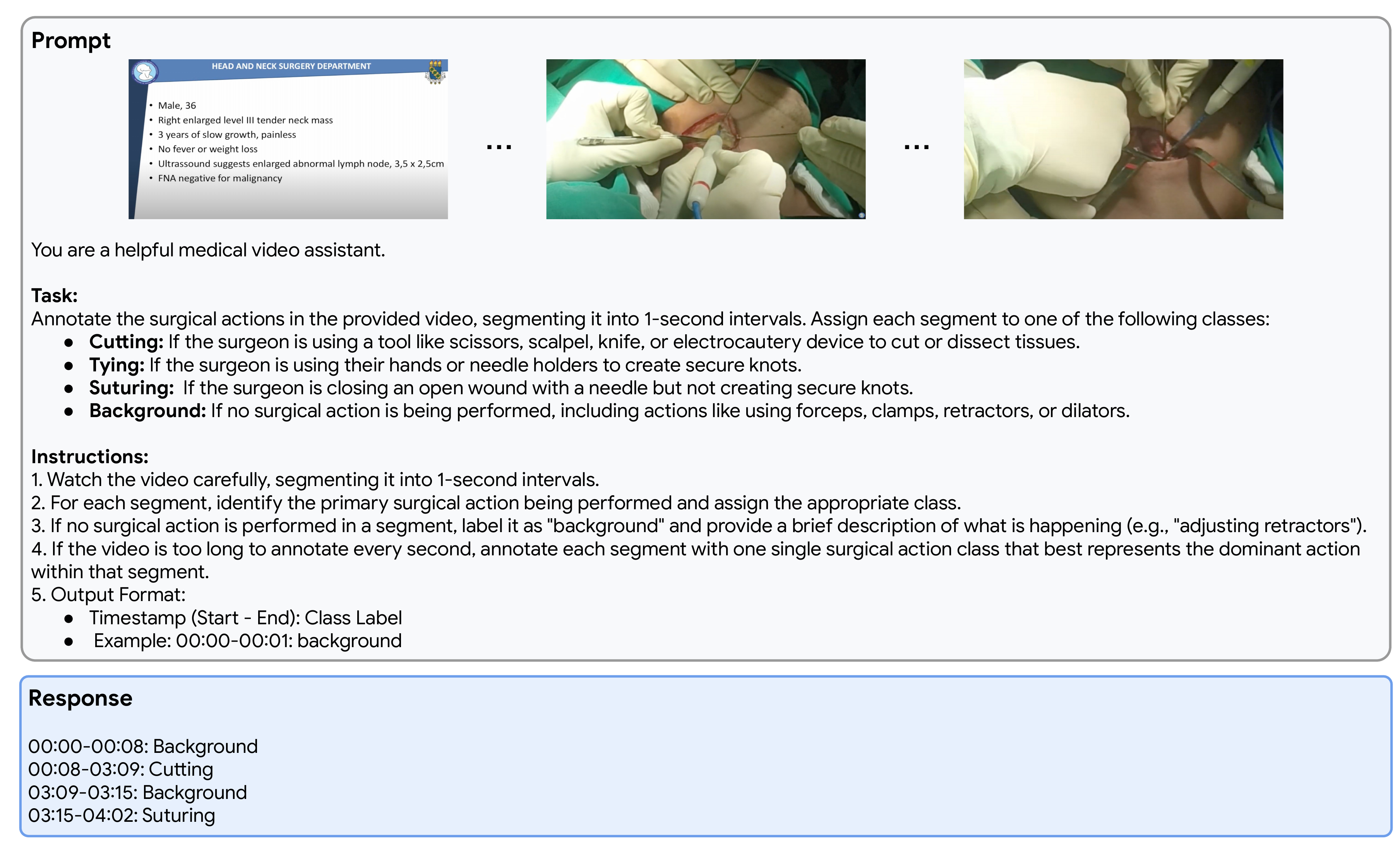}
    \caption{\footnotesize\textbf{Example of Med-Gemini-M 1.5's long-context capabilities on surgical action video span prediction.} Med-Gemini-M 1.5 analyzes a video from the Surgical Action Recognition (AVOS) dataset to annotate surgical actions in a video. It segments the video and assigns each segment a label based on the surgical action being performed (e.g., cutting, tying, suturing) or classifies it as background if no action is taking place. The ground truth AVOS annotation of this example is: 00:00-00:11: Background, 00:12-03:05: Cutting, 03:05-03:15: Background, 03:15-04:02: Suturing.}
    \label{fig:avos_ex}
\end{figure}

\begin{figure}[hbtp]
    \centering
    \includegraphics[width=\textwidth]{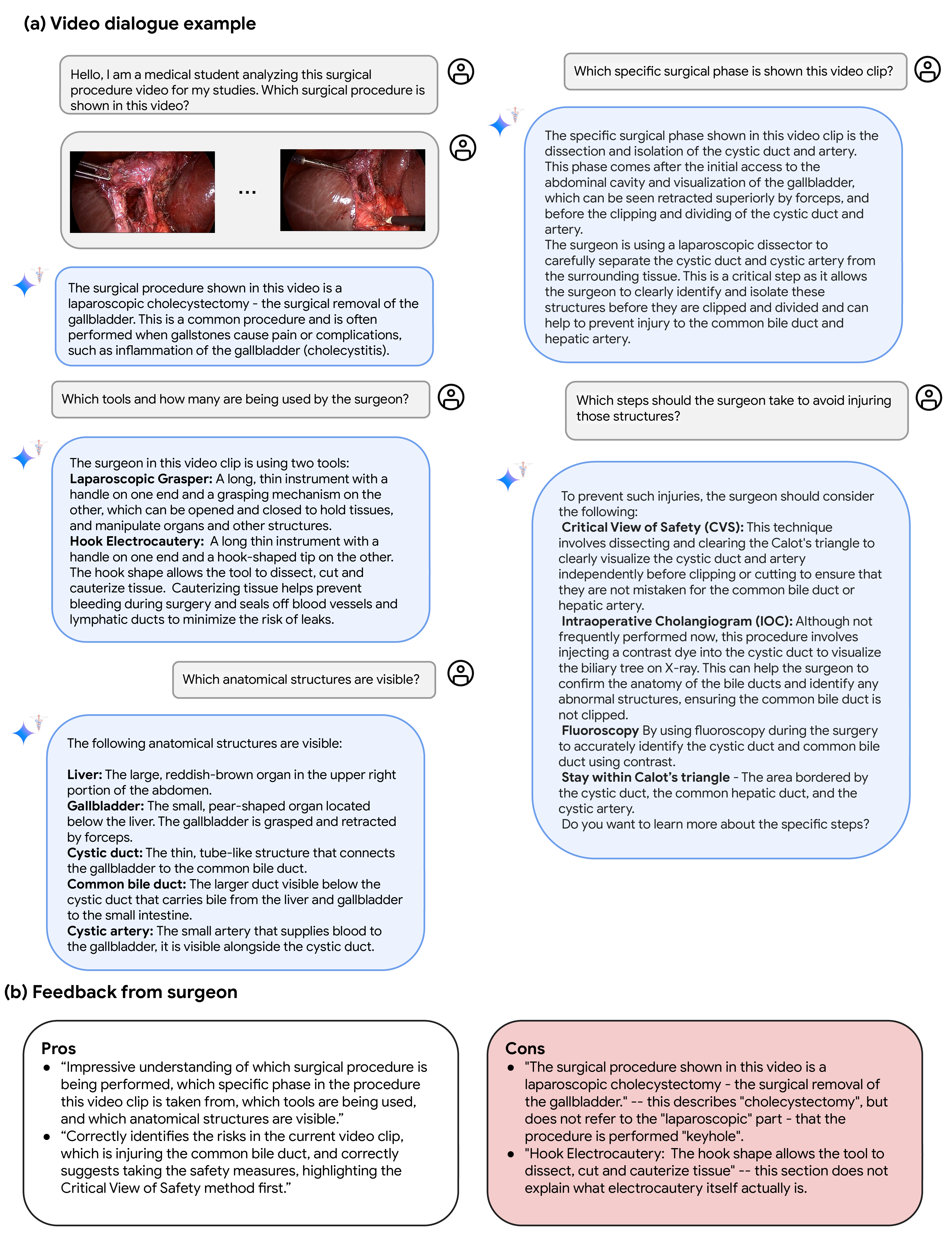}
    \caption{\footnotesize\textbf{Example of Med-Gemini-M 1.5's long-context capabilities on surgical video dialogue.} Med-Gemini-M 1.5 analyzes a video clip from the Cholec80 dataset comprising footage from a laparoscopic cholecystectomy (a keyhole operation to remove the gallbladder). The model demonstrates its ability to analyse the video and conduct a realistic dialogue with a student learning about the procedure.}
    \label{fig:cholec80_cvs_dialogue}
\end{figure}

    \paragraph{Procedural video in clinical practice and education}
    In~\Cref{fig:avos_ex}, we qualitatively preview Med-Gemini-M 1.5's ability to identify surgical actions from a video in the AVOS dataset. This ability holds potential for surgical care, promising to enhance surgical training through automated assessment, optimize operating room efficiency by analyzing workflows, and potentially guide surgeons in real-time during complex procedures for improved accuracy and patient outcomes.
    In~\Cref{fig:cholec80_cvs_dialogue}, we additionally present an example of Med-Gemini-M 1.5’s long-context capabilities on surgical video dialogue where the model analyzes a video clip comprising footage from a laparoscopic cholecystectomy. Med-Gemini-M 1.5 demonstrates its ability to analyze the video and conduct a dialogue with a student that might be learning about the procedure. These promising abilities have the potential to provide useful assistive tools for clinicians, perhaps improving patient safety or enhancing the process of medical training through educational aids or automated in-procedure assistance and guidance. The model correctly informs the user that they are observing a laparoscopic cholecystectomy and refers correctly to the key structures underlying the ``critical view of safety''. These classification tasks, if performed scalably with high accuracy, could enable better audit of procedures (for example for quality assurance), or even prospective efficiency gains from anticipation of operative stages. For more ambitious goals such as benefits to education, operative guidance or patient safety, significant further work would need to be performed to assess more nuanced and complex capabilities. For example, we did not test Med-Gemini's ability to accurately segment or highlight physical structures in the video and ground the dialogue with the relevant anatomy; or retrieve and present useful educational assets like diagrammatic representations of the displayed anatomy or guides to key operative stages. For uses such as education, pedagogical dialogue objectives would also likely be of considerable importance. Further work should explore these and other exciting new capabilities in a wider range of settings for procedural video, which is increasingly common in medicine.

\begin{figure}[hbtp]
    \centering
    \includegraphics[width=\textwidth]{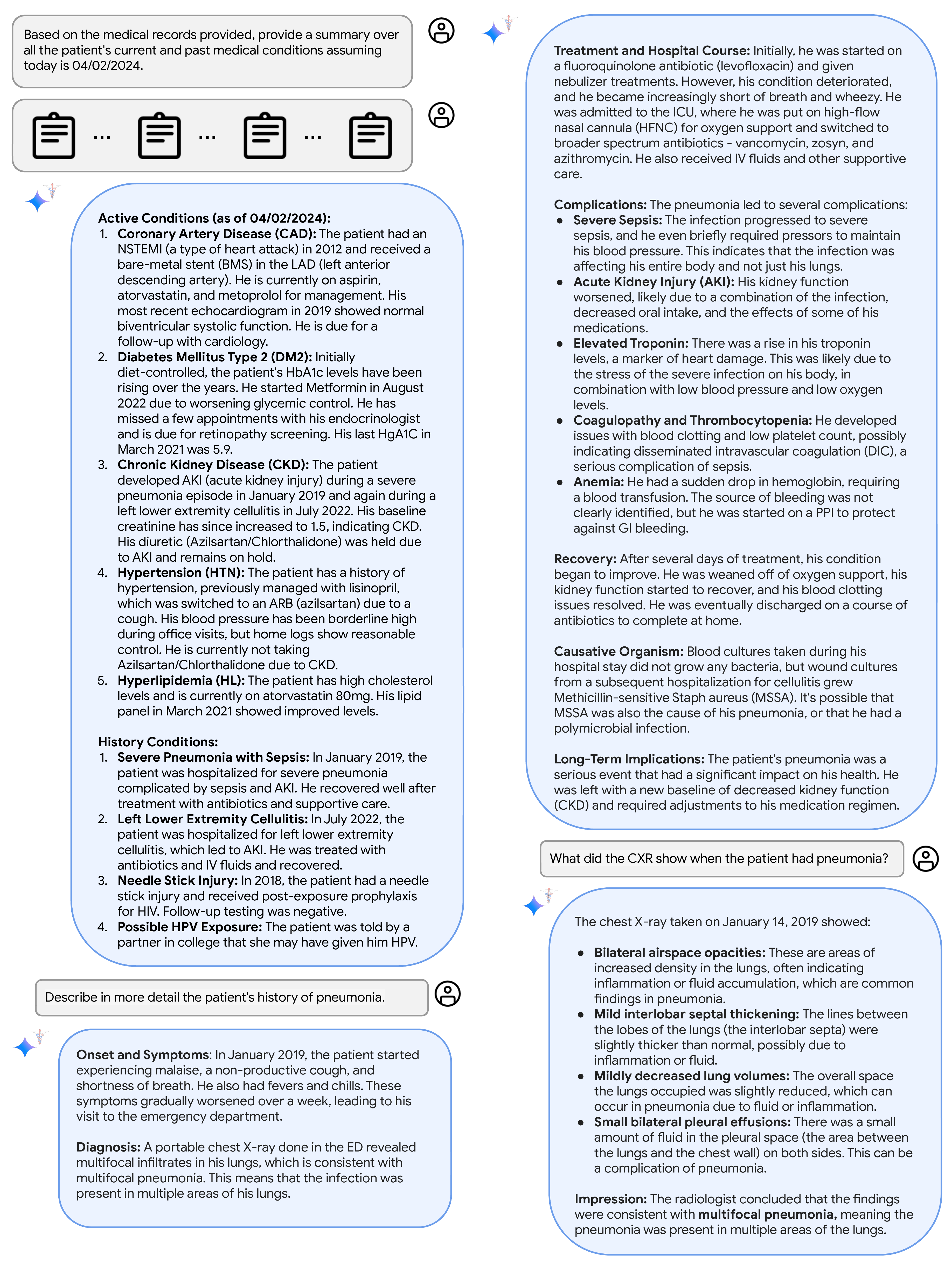}
    \caption{\footnotesize\textbf{Example of Med-Gemini-M 1.5's long-context capabilities on EHR dialogue.} Med-Gemini-M 1.5 demonstrates its ability to parse extensive medical records, providing a comprehensive summary of the patient's active and historical conditions. The conversation focuses on the patient's history of severe pneumonia, illustrating how the model's long-context understanding enables a user-friendly interface for interacting with and querying complex medical data. This capability has the potential to reduce cognitive load and improve information accessibility for both clinicians and patients.}
    \label{fig:ehr_dialogue}
\end{figure}

    \paragraph{Clinician dialogue with EHR} In~\Cref{fig:ehr_dialogue}, we demonstrate that Med-Gemini-M 1.5 effectively parses extensive medical records, synthesizing them into clear, concise summaries of active and historical conditions. Moreover, users can initiate conversations based on this summarized data, requesting more granular details from the records. Our example shows how this might include a user making natural language inquiries about specific conditions (like pneumonia) or associated diagnostic findings (such as CXR results). By streamlining access to long-form medical data and presenting the interaction in a conversational interface, this capability has the potential to significantly reduce cognitive load for clinicians and patients alike, potentially enhancing the efficiency and understanding of complex medical information without compromising staff well-being. To deliver upon this potential in real-world use would require considerable additional evaluation and research. As just one example, it would be necessary to closely examine the incidence of clinically-significant errors in retrieval or generation from grounded content; and to proactively measure and mitigate issues in dataset and model bias (as we discuss further below).
    
\begin{figure}[hbtp]
    \centering
    \includegraphics[width=\textwidth]{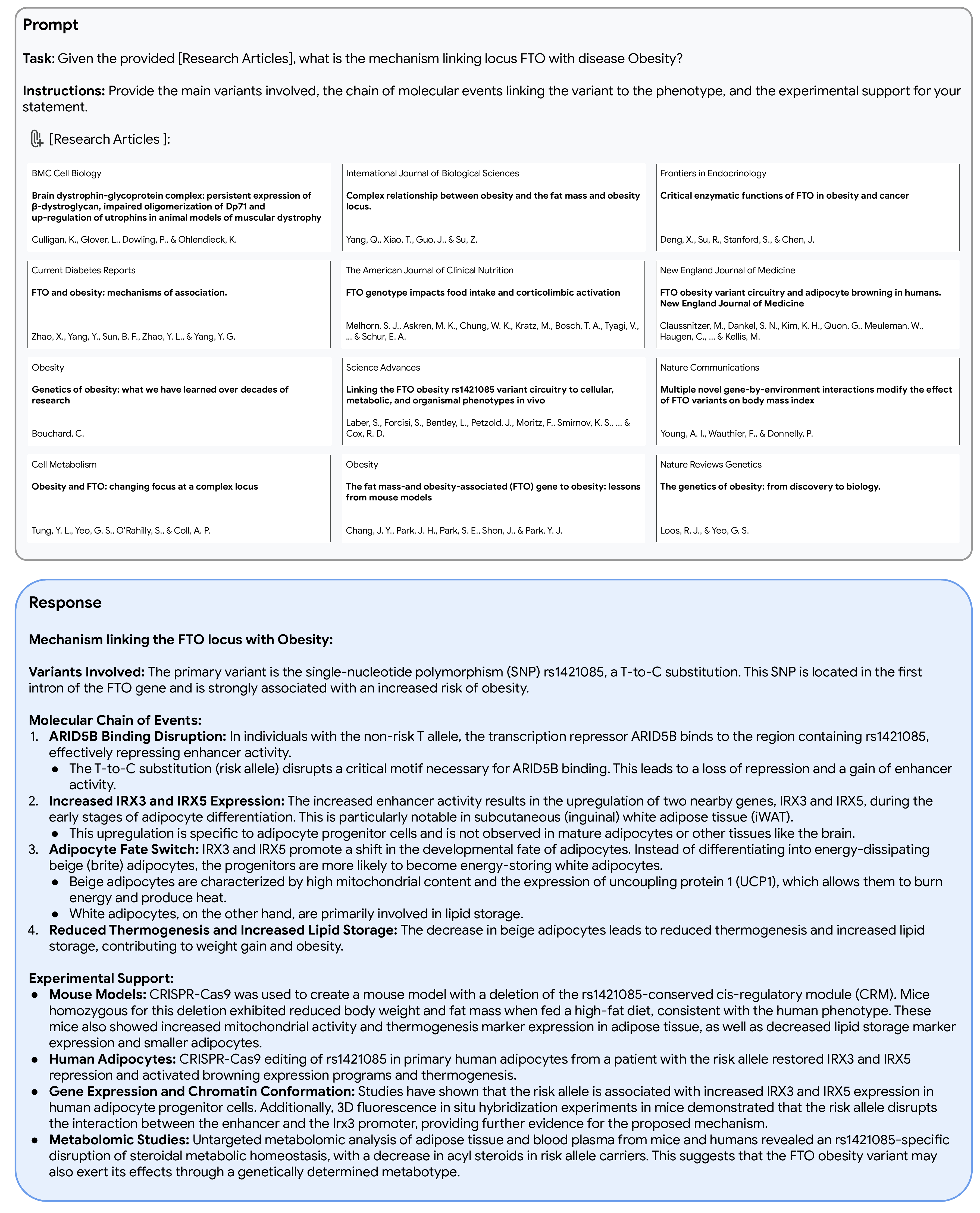}
    \caption{\footnotesize\textbf{Example of Med-Gemini-M 1.5's long-context capabilities on scientific information synthesis.} Med-Gemini-M 1.5 is given the full content of 12 research articles in portable document format (PDF) as input and is instructed to provide a succinct description of the mechanistic link between the FTO locus (a region of BMI- and obesity-associated variants within the gene \textit{FTO}) and obesity, together with a list of concrete supporting experimental results. The model generates a concise, informative, and accurate description of how the FTO locus contributes to obesity biology and presents it in a clear and digestible manner.}
    \label{fig:genomics_longcontext}
\end{figure}

    \paragraph{Biomedical research} In~\Cref{fig:genomics_longcontext}, we demonstrate Med-Gemini-M 1.5's ability to process multiple research articles concerning a specific genetic locus (FTO) and its association with obesity~\citep{loos2022genetics}. In this real-world application, Med-Gemini-M 1.5 successfully comprehends the information presented in current research (full content of 12 pre-curated research papers in portable document format) and compiles a concise summary for the user. The FTO locus we demonstrate in this example (a region of BMI- and obesity-associated variants within the gene \textit{FTO}) is a classic example of a mechanistically understood genome-wide association studies (GWAS) hit. In this exemplar, the mechanism is a relatively complex multistep process which took extensive research to pinpoint---it involves variants altering the binding of a transcriptional repressor within an intronic super-enhancer region of the \textit{FTO} gene, thereby leading to overexpression of two other genes, which ultimately promotes lipid accumulation~\citep{claussnitzer2015fto, laber2021linking}. 
    
    We evaluate Med-Gemini-M 1.5's ability to parse a large collection of academic papers on the FTO locus and provide a succinct and accessible description of the mechanistic link between FTO and obesity, together with a list of concrete supporting experimental results. As seen in~\Cref{fig:genomics_longcontext}, the model provides a concise, informative, and accurate description of how the FTO locus contributes to obesity biology and presents it in a clear and digestible manner. Improvement can be made by the model listing other well-studied variants in high linkage equilibrium with rs1421085, and by providing references of where each piece of information originated from. This example shows how Med-Gemini-M 1.5's long-context capability has clear potential to reduce cognitive load for genomic researchers and clinicians, enhancing their access to the latest findings regarding gene-disease associations; and the potential has broad relevance in other domains of biomedical and scientific research.

%% file: discussion.tex
\section{Discussion}

Med-Gemini, built upon the Gemini models, demonstrates significant advancements in clinical reasoning, multimodal understanding, and long-context processing within the medical domain. This is evidenced by its strong performance across a diverse range of 25 tasks spanning 14 medical benchmarks, encompassing medical knowledge, clinical reasoning, genomics, waveforms, medical imaging, health records and videos.

\paragraph{MedQA performance} Notably, Med-Gemini-L 1.0 achieves a new SoTA on MedQA (USMLE), a popular benchmark for medical question answering with the use of self-training based fine-tuning and search integration. Our thorough relabeling of the MedQA test set (performed by attending clinicians) reveals important insights. While MedQA (USMLE) is a useful benchmark for assessing medical knowledge and reasoning, it is essential to acknowledge its limitations. We discover that approximately 4\% of the questions contain missing information, and an additional 3\% potentially have labeling errors. Establishing definitive ground truth is frequently challenging in medicine, where inter-reader variability and ambiguity are common and medical knowledge is constantly evolving. Our observations suggest that further improvements in SoTA performance on the MedQA (USMLE) benchmark in isolation may not directly correlate to progress in the capabilities of medical LLMs for meaningful real-world tasks and as such it is important to perform more comprehensive benchmarking and evaluation representative of real-world clinical workflows~\citep{fleming2023medalign}. In general, most benchmarks have limitations around dataset size and quality. While we focus our analysis here on MedQA (USMLE), prior work has suggested similar issues with other popular benchmark datasets~\citep{xu2023elixr}. Retraining Med-Gemini-M 1.5 with a new split of the PAD-UFES-20 dermatology dataset leads to a drop of 7.1\% as compared to our results in~\Cref{tab:mm_perf}. As such, careful attention needs to be given to the size and quality of datasets when interpreting and contextualizing model performance.

\paragraph{Web search integration} Med-Gemini's integration with web search presents exciting possibilities to provide more factually accurate and reliable answers to medical queries with LLMs. In this work, we focus on training Med-Gemini-L 1.0 to issue web search queries when uncertain and integrate the results when producing responses. While the results on MedQA, NEJM CPC, and GeneTuring benchmarks are promising, significant further research is necessary. For example, we haven't considered restricting the search results to more authoritative medical sources~\citep{zakka2024almanac}, using multimodal search retrieval or performed analysis on accuracy and relevance of search results and the quality of the citations~\citep{wu2024well}. 
Further, it remains to be seen if smaller LLMs can also be taught to make use of web search. 
We leave these explorations to future work.

\paragraph{Promising multimodal conversational capabilities} The multimodal conversational capabilities of Med-Gemini-M 1.5 are promising given they are attained without any specific medical dialogue fine-tuning. Such capabilities allow for seamless and natural interactions between people, clinicians, and AI systems. As showcased in our qualitative examples, Med-Gemini-M 1.5 has the capability to engage in multi-turn clinical dialogues, request additional information such as images when needed, explain their reasoning in a comprehensible manner, and even help provide information useful for clinical decisions while appropriately deferring the final decision to human experts. This capability has significant potential for helpful real-world applications, including  assisting clinicians and patients, but of course also entails highly significant associated risks. While highlighting the potential for future research in this domain, we have not rigorously benchmarked capabilities for clinical conversation in this work as previously explored by others in dedicated research towards conversational diagnostic AI~\citep{tu2024towardsamie}. In addition, in forthcoming work, we will also rigorously explore the capabilities of Gemini in clinically specific multimodal tasks such as radiology report generation.

\paragraph{Opportunities with long-context processing} Perhaps the most notable aspect of Med-Gemini is the long-context processing capabilities because they open up new performance frontiers and novel, previously infeasible application possibilities for medical AI systems. In this work, we introduce a novel EHR task focused on identifying and verifying conditions, symptoms and procedures within very long electronic patient records. This ``needle-in-a-haystack'' retrieval task reflects a real-world challenge faced by clinicians~\citep{klerings2015information}, and Med-Gemini-M 1.5's performance demonstrates its potential to significantly reduce cognitive load and augment clinicians' capabilities by efficiently extracting and analyzing crucial information from vast amounts of patient data. The medical video question answering and annotation performance suggests these capabilities can generalize to complex multimodal data. It is worth highlighting that the demonstration of long-context capabilities is in a few-shot fashion without any task-specific fine-tuning. Such capabilities open up the possibilities of fine grained analysis and annotation of genomic and multi-omic sequence data, complex imaging modalities such as pathology or volumetric images and integrative processing with health records to uncover novel insights and assist in clinical workflows.

\paragraph{Importance of medical specialization and fine-tuning} Gemini models are inherently multimodal and have strong medical knowledge as a result of large-scale multimodal pretraining. This is reflected in impressive out-of-the-box performance on multimodal benchmarks such as NEJM Image Challenge surpassing similar generalist vision-language models such as GPT-4V by a large margin~\citep{buckley2023accuracy}. At the same time, medical knowledge and data (particularly multimodal data) is unique and complex and unlikely to be seen on the public internet commonly used to train LLMs. Gemini is a strong intelligence substrate but further fine-tuning, specialization and alignment of even such powerful models are necessary before use in the medical domain. At the same time, given the general capabilities of Gemini, the amount of data needed for such specialization and alignment is much lower than prior generation of medical AI systems~\citep{azizi2023robust} and it is indeed possible to efficiently adapt such models even to previously unseen but important medical modalities such as ECGs with relative efficiency as demonstrated here.

\paragraph{Need for rigorous evaluation beyond benchmarks}
To the best of our knowledge, this work is the most comprehensive evaluation of medical LLMs and LMMs. The work includes evidence of new capabilities for medical AI and tasks that suggest real-world utility. This is particularly reinforced by strong performance of our models in evaluations of medical summarization and referral note generation. Diagnostic tasks draw considerable attention in research, but carry significant regulatory, clinical and equity-related risks that require addressing before real-world implementation is safe and feasible. The more common real-world use cases of generative AI in healthcare are therefore in non-diagnostic tasks, where errors have a lower risk-profile yet model outputs can significantly improve the efficiency of care providers by alleviating administrative burdens and assisting complex information retrieval or synthesis required in day-to-day work. At the same time, even for such non-diagnostic tasks, assurance of real-world impact requires evaluation grounded in specific use-cases and environments. These evaluations lie beyond the scope of initial benchmarking, and our results should be interpreted with appropriate caution. To assess downstream consequence and generalization of the promise we demonstrate here to real-world clinical workflows, practitioners should adhere to best practices of responsible AI, rigorously measuring multiple endpoints including equity~\citep{pfohl2024toolbox}, fairness and safety in the intended environment while also considering the multiple socio-technical factors that are use-case specific determinants of impact. Finally, it is worth noting that while we have considered 14 diverse and challenging benchmarks in this study, over 350 medical benchmarks are available in the community~\citep{paperswithcode_medical}.

\paragraph{Responsible AI}
Our work has been primarily focused on capabilities and improvements and the art of the possible with Gemini models. An important focal area for future exploration is the integration of the responsible AI principles throughout the model development process~\citep{pfohl2024toolbox}, including, but not limited to, the principles of fairness, privacy, equity, transparency and accountability. Privacy considerations in particular need to be rooted in existing healthcare policies and regulations governing and safeguarding patient information. Fairness is another area that may require attention, as there is a risk that AI systems in healthcare may unintentionally reflect or amplify historical biases and inequities~\citep{char2018implementing, obermeyer2019dissecting, cirillo2020sex, gichoya2022ai, abramoff2023considerations, pfohl2024toolbox}, potentially leading to disparate model performance and harmful outcomes for marginalised groups. Such health disparities have been identified across gender~\citep{kent2012gender}, race~\citep{williams2015racial, obermeyer2019dissecting}, ethnicity~\citep{razai2021mitigating}, socioeconomic status~\citep{steptoe2020lower}, sexual orientation~\citep{medina2021health}, age~\citep{jackson2019associations}, and other sensitive and/or protected personal characteristics. There is an increasing need for a deep intersectional analysis of impact~\citep{iyer2008intersections, lopez2017health}, though this remains a hard technical problem~\citep{cabrera2019fairvis, yang2020fairness, wang2022towards}, and an active area of research. 

As we demonstrate new capabilities for LLMs and LMMs, new opportunities arise for potential issues at the confluence of dataset bias~\citep{ganapathi2022tackling}, model bias~\citep{liu2023translational}, and the socio-technical considerations for individual use cases. In the context of the capabilities we have discussed, these issues may potentially occur in in-context learning within the long-context utilization of potentially biased examples and instructions, in search integration, the dynamics of self-training, or multimodal understanding with fine-tuning and customized data encoders. Within each of these capabilities, there could be multiple points at which such biases may need to be considered. When it comes to web search integration, biases could come up at query construction time, get reflected in the returned result set~\citep{novin2017making}, or be embedded within each of the linked external sources, and manifest in various other subtle ways, e.g. how the results are integrated into the generative reasoning process when producing the final answer. With multimodal models, biases may occur in each of the individual modalities separately, or only be apparent jointly, across co-dependent modalities of the data~\citep{srinivasan2021worst, mandal2023multimodal}. A comprehensive analysis of potential issues may need to consider each of these points separately, but also holistically as they are all parts of a complex system. These systems may also need to be thoroughly evaluated not only in isolation, but also with human experts in the loop.

However, these new capabilities also present an opportunity to mitigate prior issues and dramatically improve accessibility across use-cases. For example, new long-context capabilities in medicine may enable a model's users to solve complex problems at inference time without the need for engaging in model fine-tuning, as the data can be utilized directly within the context of the query, followed by a set of natural language instructions. Previously, users of such systems would have needed to possess engineering expertise and invest additional time and resources in fine-tuning custom models for tackling such complex tasks. Web search integration, on the other hand, may prove to be invaluable when it comes to rapidly integrating newly developed pieces of medical knowledge and external consensus on what is a highly dynamic and non-stationary medical landscape. The COVID-19 pandemic has shown just how quickly the public health understanding and recommendations may need to get updated, and it also highlighted the overall danger posed by medical misinformation~\citep{kouzy2020coronavirus}. Models that can reliably consume reputable up-to-date external sources may be far less likely to lead to such misinformation. Similar new opportunities are presented by the other model capabilities, though further study is needed to develop a robust evaluation framework to assess the associated risk of bias and unfair outputs (whether individually or jointly across complex use-cases), with such assessments sociotechnically grounded in real settings for specific clinical use-cases.

\section{Conclusion}
Large multimodal language models are ushering in a new era of possibilities for health and medicine. The capabilities demonstrated by Gemini and Med-Gemini suggest a significant leap forward in the depth and breadth of opportunities to accelerate biomedical discoveries and assist in healthcare delivery and experiences. However, it is paramount that advancements in model capabilities are accompanied by meticulous attention to the reliability and safety of these systems. By prioritizing both aspects, we can responsibly envision a future where the capabilities of AI systems are meaningful and safe accelerators of both scientific progress and care in medicine.

%% file: appendix.tex
\section*{Appendix}
\setcounter{table}{0}
\setcounter{figure}{0}
\renewcommand{\thetable}{A\arabic{table}}
\renewcommand{\thefigure}{A\arabic{figure}}

\section{Supplementary Table for Figure 1}

\begin{table}[ht!]
\footnotesize
\centering
\resizebox{1.0\textwidth}{!}{
\begin{tabular}{ccccccc}
\toprule
\textbf{Capability} & \textbf{Task} & \textbf{Metric} & \textbf{Med-Gemini} & \textbf{Previous SoTA} & \textbf{Best GPT-4 Method} & \textbf{SoTA Reference} \\
\midrule
Advanced Text Reasoning & NEJM CPC & Top-10 Accuracy & \textbf{72.3} & 59.1 & 50.0 & \cite{mcduff2023towards} \\
 & GeneTuring & Averaged accuracy & \textbf{53.3} & 48.6 & 48.6 & \cite{hou2023geneturing} \\
 & MedQA & Accuracy & \textbf{91.1} & 90.2 & 90.2 & \cite{nori2023can} \\
\midrule
Multimodal Understanding & NEJM Image & Accuracy & \textbf{69.7} & 61.0 & 61.0 & \cite{buckley2023accuracy} \\
 & USMLE-MM & Accuracy & \textbf{93.5} & 80.4 & 80.4 & Reproduced \\
 & ECG-QA & Accuracy & \textbf{57.7} & 51.6 & 51.6 & \cite{oh2023ecgqa} \\
 & MMMU-HM & Accuracy & \textbf{67.3} & 64.7 & 64.7 & \cite{yue2023mmmu} \\
 & Path-VQA & Token F1 & \textbf{64.7} & 62.7 & 36.0 (Reproduced) & \cite{tu2024towardsmpm} \\
 & PAD-UFES-20 & Accuracy & 85.9 & \textbf{88.0} & 50.0 (Reproduced) & \cite{tu2024towardsmpm} \\
 & Slake-VQA & Token F1 & 87.5 & \textbf{89.3} & 41.0 (Reproduced) & \cite{tu2024towardsmpm} \\
\midrule
Long-context Processing & MedVidQA & mIoU & \textbf{43.4} & 27.5 & N/A & \cite{li2022towards} \\
 & MedVidQA w/ subtitles & mIoU & \textbf{65.8} & 58.3 & N/A & \cite{weng2023visual} \\
 & Long EHR & F1 & 0.77 & \textbf{0.78} & N/A & \cite{feder2022building} \\
& Surgery Video CVS Assessment & Accuracy & 50.0 & \textbf{67.0} & 0.290 & Reproduced \\

\bottomrule
\end{tabular}
}
\caption{\footnotesize\textbf{Performance results of bar plot in Figure 1.} We display our aggregated results comparing Med-Gemini to the previous state-of-the-art (SoTA) and the best GPT-4 methods across text-based, multimodal, and long-context tasks. For benchmarks where we could not find GPT-4 (or GPT-4V) reported numbers in literature, we run evaluations on the same test sets using public APIs for a head-to-head comparison, using same few-shot prompts as the corresponding Med-Gemini model including instructions to ensure outputs are correctly formatted. Note that GPT-4 results are not available (N/A) for three long context tasks due to limitations of the context window of the public GPT-4 / GPT-4V APIs}
\label{tab:hero_table}
\end{table}

\input{related}

\setcounter{table}{0}
\setcounter{figure}{0}
\renewcommand{\thetable}{C\arabic{table}}
\renewcommand{\thefigure}{C\arabic{figure}}
\section{Additional details on advanced reasoning text-based tasks}

\subsection{Text-based fine-tuning \& evaluation datasets}

\begin{table}[H]
\footnotesize
\centering
\resizebox{1.0\textwidth}{!}{
\begin{tabular}{ccccc}
\toprule
\textbf{Task type} & \textbf{Datasets} & \multicolumn{1}{l}{Sample size} & \textbf{Description} & \textbf{Reference} \\
\midrule
Multiple-choice question answering & MedQA & 10177 & Multiple-choice questions from MedQA & \cite{jin2021disease} \\
\begin{tabular}[c]{@{}c@{}}Multiple-choice question answering\\ with reasoning\end{tabular} & MedQA-R, MedQA-RS & 20354 & \begin{tabular}[c]{@{}c@{}}Multiple-choice questions from MedQA with \\ synthetically generated reasoning examples\end{tabular} & Novel \\
\midrule
Long-form question answering & HealthSearchQA, LiveQA, MedicationQA & 260 & Clinician-written long-form responses & \cite{singhal2023large} \\
\midrule
Summarization & MIMIC-summaries & 65 & Clinician-written summaries of medical notes & \cite{tu2024towardsamie} \\ 
\bottomrule
\end{tabular}
}
\caption{\footnotesize\textbf{Overview of datasets used for text-based instruction fine-tuning.} The dataset mixture and synthetic data are curated to improve Med-Gemini-L 1.0's reasoning and ability to make use of web search.}
\label{tab:text_ft}
\end{table}

\begin{table}[H]
\footnotesize
\centering
\resizebox{1.0\textwidth}{!}{
\begin{tabular}{cccccc}
\toprule
Task Type & Modality & Dataset & Test sample size & Description & Reference \\
\midrule
Close-ended QA & Text & MedQA & 1273 & US medical licensing exam-style, multiple-choice & \cite{jin2021disease} \\
Open-ended QA & Text & NEJM CPC & 303 & Complex diagnostic challenging in NEJM & \cite{mcduff2023towards} \\
Open/Close-ended QA & Text & GeneTuring & 600 & Commonly seen tasks in genomics research & \cite{hou2023geneturing} \\
Long-form generation & Text & Clinical Abstraction & 81 & Meaningful summarization in clinical practice and research  & \Cref{app:long_form_sxs} \\
\bottomrule
\end{tabular}
}
\caption{\footnotesize\textbf{Overview of the evaluation benchmarks used for text-based reasoning tasks.}}
\label{tab:text_eval}
\end{table}

\subsection{MedQA (USMLE) Relabeling}
\label{app:medqa-relabeling}

The main objectives of this rater study are to identify (a) unanswerable questions due to missing information, (b) potential label errors, and (c) potentially ambiguous questions \citep{stutz2023evaluating}. To this end, we carefully design a two-step study as follows:

\begin{itemize}
    \item \textbf{Step 1:} Given the MedQA (USMLE) question and all four answer options:
    \begin{itemize}
        \item (Q1) We ask ``Are any of the options appropriate to answer this question?''
        \item (Q2) If yes, ``Select one or more options to answer the question.'' (Multi-select)
        \item (Q3) We ask ``Is there any additional information (such as figures, plots, lab results, or similar) referenced in the question that is missing?''
        \item (Q4) If yes, we ask ``Do you think having access to the missing information would change your answer?''
    \end{itemize}
    \item \textbf{Step 2:} After the rater completes step 1, they are presented with the ground truth answer from MedQA:
    \begin{itemize}
        \item (Q1) We ask ``Having revealed the question bank’s answer key, does your answer from before change?''
        \item (Q2) If yes, we repeat the first two questions from above.
    \end{itemize}
\end{itemize}

A key consideration that leads to this \textbf{two-step approach} is to reveal the MedQA (USMLE) ground truth at the right time to avoid biasing the rater with the ground truth when answering questions about potentially missing information in the question (Q3 and Q4). For properly identifying label errors, however, we present the raters with the ground truth so they can decide to disagree (Q1 and Q2 in step 2). To identify potentially ambiguous questions (allowing multiple ``good'' or true answers), we further allow raters to select multiple options as answers\footnote{The additional question asking whether any option is appropriate (before revealing the multi-select) is due to technical constraints.}. When asking about potentially missing information, we aim to identify whether this missing information is critical to answer the question.

\begin{table*}[ht!]
\centering
\scriptsize
\begin{tabular}{ccccc}
\toprule
\multicolumn{5}{c}{\textbf{Rater agreement in \% against vote}} \\
\midrule
 &  & All & Med-Gemini incorrect & Med-Gemini correct \\
\midrule
info missing & majority & 95.6 & 95.6 & 95.6 \\
info missing & unanimous & 94.0 & 94.4 & 94.0 \\
label errors & majority & 89.6 & 80.8 & 90.5 \\
label errors & unanimous & 87.6 & 74.6 & 88.8 \\
ambiguous & majority & 94.9 & 92.9 & 95.0 \\
ambiguous & unanimous & 94.6 & 92.3 & 94.9 \\
\bottomrule
\toprule
\multicolumn{5}{c}{\textbf{Rater agreement of answers as average overlap in \%}} \\
\midrule
 &  & All & Med-Gemini incorrect & Med-Gemini correct \\
\midrule
before & majority & 49.9 & 36.4 & 51.2 \\
before & unanimous & 49.1 & 36.0 & 50.4 \\
after & majority & 75.9 & 54.8 & 77.9 \\
after & unanimous & 74.6 & 53.9 & 76.6 \\
\bottomrule
\end{tabular}
\caption{\footnotesize\textbf{Annotation agreement.} \textbf{Top:} Agreement of individual ratings against the majority or unanimous vote for various rating tasks of interest. \textbf{Bottom:} Agreement of raters' answers in terms of average overlap before and after having revealed the MedQA ground truth.}
\label{tab:app-medqa}
\end{table*}

We recruit a total of 18 primary care physicians (PCPs) from the US to participate in the study. We select PCPs located in the US because MedQA comprises USMLE-style questions across many specialties. For each MedQA (USMLE) question, we collect at least three ratings from independent raters. While the original MedQA work in~\citep{jin2021disease} evaluated expert performance with access to additional text material, our raters are not instructed to use any material. However, we do not explicitly control for this. PCPs take an average of 255 seconds to complete one question; 98\% take less than 10 minutes.

For each question, we aggregate the ratings in order to identify, e.g., label errors with high certainty. First, we evaluate the agreement of each rater against the majority or unanimous vote in~\Cref{tab:app-medqa}. Specifically, we consider the rater agreement for four rating tasks of interest: whether information is missing, whether there is a label error (i.e., the rater's answer after revealing the MedQA (USMLE) ground truth does \emph{not} include the ground truth answer from MedQA (USMLE)), whether a question is ambiguous (i.e., the rater's answer includes more than one option even after revealing the MedQA (USMLE) ground truth) and agreement between raw answer options selected in terms of average overlap (each rater can select none or multiple options). For the former three, agreement is generally high (>87\%), although it is usually lower on questions where Med-Gemini-L 1.0 makes mistakes. For the third, in contrast, agreement in terms of average overlap between all pairs of answers is significantly lower: typically around 75\% when raters have seen the MedQA (USMLE) ground truth, but agreement drops to around 50\% if the ground truth is not revealed to raters.

\begin{figure*}[ht!]
\begin{minipage}{0.48\textwidth}
\includegraphics[width=\textwidth]{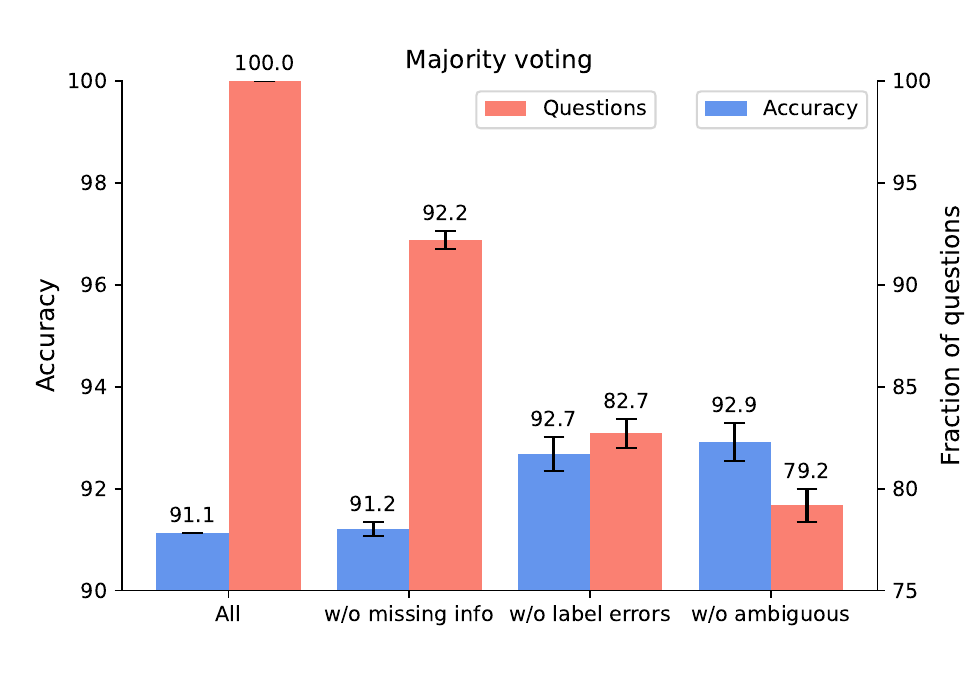}
\end{minipage}
\begin{minipage}{0.48\textwidth}
\includegraphics[width=\textwidth]{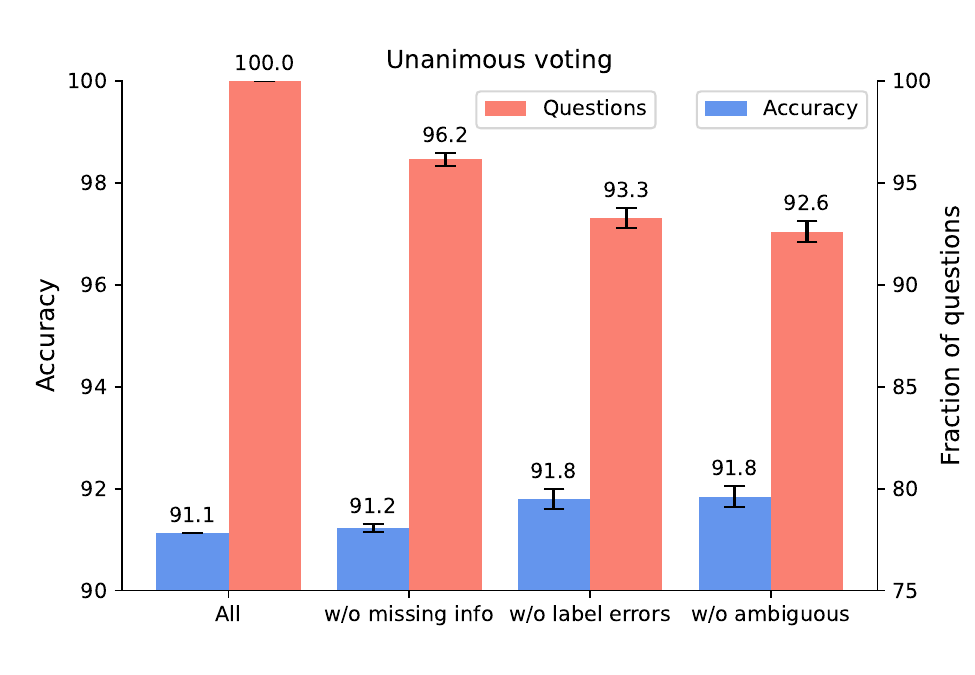}
\end{minipage}
\caption{\footnotesize\textbf{MedQA (USMLE) results after re-annotation.} Results complementary to Figure \ref{fig:text_results}b showing accuracy (blue) and remaining MedQA (USMLE) questions (red) after filtering questions with missing information, label errors or questions deemed ambiguous when aggregating ratings using majority voting (left) or unanimous voting (right).}
\label{fig:app-medqa-uncertainty}
\end{figure*}

To measure the impact of filtering MedQA (USMLE) questions with missing information or label errors on evaluation while taking into account annotation uncertainty, we perform a bootstrapping experiment. Specifically, we repeatedly sample a committee of three raters per question (with replacement). For each committee of raters, we perform majority or unanimous voting to identify questions with missing information or label errors to be filtered for evaluation. This can be seen as an instance of the evaluation framework proposed in \citep{stutz2023evaluating}. The advantage of bootstrapping over simple voting is that we get reliable uncertainty estimates that make sure we can identify performance changes as statistically significant. We repeat this experiment $1000$ times and report average and standard deviation of accuracy and fraction of remaining questions in Figure \ref{fig:app-medqa-uncertainty}.

While questions with missing information or label errors can be identified with high confidence due to high agreement, it is more difficult to judge whether a question is \emph{ambiguous}. Here, we define a question as being ambiguous if it allows for multiple answer options to be correct. Most questions in the MedQA (USMLE) test set specifically ask for the ``best'', ``most likely'' or ``most appropriate'' option. However, it is largely unclear whether answers do indeed only allow for one option to be e.g. the ``best next step in management'' of a case. After excluding questions with missing information and label errors using majority voting, raters selected on average 1.065 options, indicating that some questions might indeed be ambiguous. This increases to 1.119 after revealing the ground truth. To take this into account during evaluation, we define a rating as ambiguous if the rater selects more than one option after revealing the ground truth. We then follow the same analysis as above and show results in~\Cref{fig:app-medqa-uncertainty}. Overall, we find that filtering label errors has the biggest impact on Med-Gemini-L 1.0's performance, while filtering for missing information or ambiguous questions can reduce the number of questions but does not change accuracy significantly.

\subsection{Additional results on NEJM clinico-pathological conference dataset} \label{app:cpc}

We present Top-1 and Top-10 performance on the NEJM clinical pathology case studies as broken down by the primary speciality of the cases (as identified by NEJM) in Table~\ref{tab:nejm_specialties}, for all specialties with at least 10 cases. In most specialties, Internal Medicine, Pediatrics, and Psychiatry, the best Top-1 and Top-10 performance is achieved by either Med-Gemini-L 1.0 without search or the model with search. 

\begin{table}[ht!]
\footnotesize
\centering
\resizebox{1.0\textwidth}{!}{
\begin{tabular}{lcccccccccc}
\toprule[1.5pt]
& \multicolumn{2}{c}{Previous SoTA} & \multicolumn{4}{c}{Clinician} & \multicolumn{4}{c}{Med-Gemini} \\
& \multicolumn{2}{c}{AMIE} & \multicolumn{2}{c}{Without Search} & \multicolumn{2}{c}{with Search} & \multicolumn{2}{c}{Without Search} & \multicolumn{2}{c}{With Search} \\
\cmidrule(lr){2-3} \cmidrule(lr){4-5} \cmidrule(lr){6-7} \cmidrule(lr){8-9}  \cmidrule(lr){10-11} 
Metrics & Top-1$^\uparrow$ & Top-10$^\uparrow$ & Top-1$^\uparrow$ & Top-10$^\uparrow$ & Top-1$^\uparrow$ & Top-10$^\uparrow$ & Top-1$^\uparrow$ & Top-10$^\uparrow$  & Top-1$^\uparrow$ & Top-10$^\uparrow$  \\ \midrule
Internal Med (159 Cases) & 27.7\% & 61.6\% & 15.5\% & 34.6\% & 24.5\% & 47.8\% & 24.5\% & 64.8\% & 31.4\% & 74.8\% \\[1.2pt]
Neurology (42 Cases) & 26.8\% & 56.1\% & 17.1\% & 31.7\% & 22.0\% & 36.6\% & 31.0\% & 66.7\% & 26.2\% & 61.9\%  \\[1.2pt]
Pediatrics (33 Cases) & 30.3\% & 45.5\% & 6.1\% & 22.7\% & 12.1\% & 33.3\% & 21.2\% & 45.5\% & 12.1\% & 48.5\% \\[1.2pt]
Psychiatry (10 Cases) & 50.0\% & 70.0\% & 20.0\% & 50.0\% & 20.0\% & 60.0\% &  50.0\% & 100.0\% & 60.0\% & 90.0\% \\[1.2pt]
\bottomrule[1.5pt]
\end{tabular}
}
\caption{\footnotesize\textbf{Breakdown of performance on NEJM case studies by speciality.} Performance is reported for the specialities with at least 10 cases.}
\vspace{-0.2cm}
\label{tab:nejm_specialties}
\end{table}

\subsection{Real-world use cases for advanced reasoning on text-based tasks}\label{app:long_form_sxs}

We instruction fine-tune and evaluate Med-Gemini-M 1.0 on three challenging real-world tasks requiring long-form text generation. Summary results are shown in~\Cref{fig:long_form_sxs}. Detailed results of additional evaluation axes are shown in~\Cref{tab:long_form_sxs}. Datasets and evaluation procedures for each task are described in more detail below.

\begin{table}[ht!]
\footnotesize
\centering
\resizebox{1.0\textwidth}{!}{
\begin{tabular}{cccccccc}
\toprule
\textbf{Task} & \textbf{Num. Examples} & \textbf{Evaluation-Axis} & \textbf{Expert Preferred} & \textbf{Tied} & \textbf{Med-Gemini} & \textbf{p-value} \\
\midrule
After-Visit Summary & 31 & Accuracy & 19\% & 55\% & 26\% & \textbf{p < 0.001}   \\
&  & Coverage & 48\% & 16\% & 35\% &  p = 0.861 \\
&  & Succinctness & 29\% & 10\% & 61\% & \textbf{p = 0.017} \\
&  & Coherence & 29\% & 13\% & 58\% &  \textbf{p = 0.017} \\
&  & Overall & 32\% & 3\% & 65\% &  \textbf{p = 0.046} \\
\midrule
Referral Letter & 25 & Overall & 0\% & 8\% & 92\% & \textbf{p < 0.001} \\
\midrule
Cochrane Simplification & 25 & Accuracy & 52\% & 8\% & 40\% & p = 0.846 \\
&  & Coverage & 12\% & 12\% & 76\% & \textbf{p < 0.001} \\
&  & Succinctness & 4\% & 4\% & 92\% & \textbf{p < 0.001} \\
&  & Reading Level & 0\% & 0\% & 100\% & \textbf{p < 0.001} \\
&  & Overall & 12\% & 0\% & 88\% & \textbf{p < 0.001}\\
\bottomrule
\end{tabular}
}
\caption{\footnotesize\textbf{Evaluation of Med-Gemini on long-form text-based tasks via side-by-side comparison with experts.} Tasks include generation of after-visit summaries, referral letters, and simplified summaries of systematic biomedical reviews. Evaluation is performed by clinician raters. P-values are for whether the rate at which Med-Gemini-M 1.0 is preferred or tied with experts is $0.5$ (two-sided t-test).}
\label{tab:long_form_sxs}
\end{table}

\paragraph{Medical summarization evaluation}
This task involves generating an after-visit summary (AVS) from a de-identified history and physical (H\&P) note. An H\&P note is a detailed document in which a healthcare provider records the findings of a patient visit, including a patient's health background, their current symptoms, and the findings of a physical examination. It is largely written for other \textbf{healthcare providers} to ensure coordinated care. An AVS, on the other hand, is a structured report that \textbf{patients} receive at the end of a medical visit summarizing the most important aspect of the visit and their health status.

A set of 31 de-identified H\&P notes is sampled from a dataset of de-identified medical notes from outpatient visits to family medicine or internal medicine providers. The expert after-visit summaries are written by U.S based clinicians following guidelines based on \citep{sieferd2019after}, and further refined by a second round of clinicians to further increase quality.

Med-Gemini-M 1.0 is prompted to generate an after-visit summary given the de-identified H\&P note as follows:
\begin{tcolorbox} \small
Please read through the provided medical note describing an outpatient visit and extract the relevant information for each of the following 12 fields: \\

- Patient name/age/gender: This should summarize the patient’s name, age and gender. It should use the format: “[Patient name], [age] year old [gender]”. If the name is not mentioned in the note, please answer "Not available". \\
- Today I was seen by: This field should provide the name of the provider. If the provider seen for the note being summarized is not mentioned, please answer “Not available”. \\
- I came in today for: This field should indicate the chief complaint or complaints that caused the visit. \\
- New health issues identified today are: This field should indicate any new diagnoses or other issues identified as a result of the visit being summarized. If the issue is a pre-existing condition identified in the past, please answer "No new diagnosis". \\
- Other health issues I have are: This field should indicate any pre-existing health issues identified in notes. \\
- Today we accomplished: This field should summarize the main topics of discussion and results of any procedures performed during the current visit. The summary could be a short list of procedures, or could be a text description of the patient’s experience. Please be as brief as possible when providing details, such as test results or medication names. Describing the experience from the patient’s point of view, using phrases like “my visit”, “my condition”. \\
- My important numbers: This field should provide the results of any measurements relevant to the  visit, including vitals. Provide the results of any numeric measurements relevant to the visit, including vitals, laboratory studies, or pain scores. Please include the numbers that should be monitored. Do not fabricate numbers that are not presented in the note. \\
- Changes to my medications are: This field should specify any medications that were added, for which the doses were updated, or which are no longer needed after the visit. Please specify both newly added and stopped medications when possible. If no changes are apparent from the note, please answer “no changes”. \\
- Other medications I have are: If the note indicates any existing medications for the patient that the patient should continue taking without changes, list them here. If no medications are indicated in the note, please  “Not specified”. \\
- My next steps are: This field should document the patient’s next steps, including any actions they should take, test results they should expect, and follow-up visits they should schedule, along with the appropriate time frames for each. \\
- I should seek immediate medical attention if: If the note specifies any conditions for which the patient should immediately seek care, specify it here. Be sure to only include conditions that are mentioned in the note. If no conditions are mentioned, write "Not specified". \\
- Other comments from my provider: This is an optional extra field that captures any additional relevant information the provider indicated in the notes that it would be useful for the patient to know. Do not include information that is already listed in the previous field. \\ \\
For each field, write at a sixth-grade reading level and avoid using abbreviations or jargon. \\

Note: \{MEDICAL\_NOTE\}\\
After Visit Summary:
\end{tcolorbox}

Physician raters are presented with the H\&P note, the clinician generated AVS and our model's generated AVS. Each example is evaluated once by one of three different U.S.-based physicians across the following axes:

\begin{tcolorbox} \small
\textbf{Accuracy:} \\
Which summary is more accurate? (Are all statements in the summary correct?)\\
- A - B - Tie\\
\textbf{Coverage:} \\
Which summary has better coverage? (Does it include all relevant aspects of the note?) \\
- A - B - Tie\\
\textbf{Coherence:} \\
Which summary is easier to read? (Is the summary comprehensible to a consumer with no specific medical knowledge at a 6th-grade reading level?) \\
- A - B - Tie\\
\textbf{Succinctness:} \\
Which summary is more succinct? (Is the summary longer than it needs to be?) \\
- A - B - Tie\\
\textbf{Overall:} \\
Which summary feels higher quality to you? (Beyond these metrics, is there a gut feeling about the quality of the summary?) \\
- A - B - Tie\\
\end{tcolorbox}

\paragraph{Referral letter generation evaluation}
This task involves generating a referral letter to another healthcare provider given a de-identified outpatient medical note that contains a recommendation for a referral. A medical referral letter is a formal document written by a healthcare professional that requests another healthcare professional to evaluate or treat a patient. It serves as a communication tool between healthcare providers, ensuring continuity of care and facilitating appropriate treatment for the patient.

A set of de-identified medical notes requiring inter-specialty evaluation are manually selected by clinicians from a de-identified electronic healthcare record dataset. They then generate referral letters, which are further reviewed for quality by a U.S. board-certified clinician.

Med-Gemini-M 1.0 is prompted to generate a referral letter given the medical note as follows:
\begin{tcolorbox} \small

You will be provided with a medical note describing a patient visit. The medical note will contain a recommendation that the patient be referred to another healthcare provider. Your task is to generate the medical referral letter for this healthcare provider.\\\\
A medical referral letter is a formal document written by a healthcare professional that requests another healthcare professional to evaluate or treat a patient. It serves as a communication tool between healthcare providers, ensuring continuity of care and facilitating appropriate treatment for the patient.\\\\
Medical Note: \{MEDICAL\_NOTE\}\\
Referral Letter:
\end{tcolorbox}

Physician raters are presented with the outpatient note, the clinician generated referral letter and our model's generated referral letter. They are blinded to the source of each referral letter and asked to perform the following comparison:

\begin{tcolorbox} \small
\textbf{Instructions:} \\
You are given a medical note that mentions a referral to another healthcare provider. Imagine you need to write a referral letter based on the information in the note. You are provided with draft referral letters written by two different assistants. Which draft do you prefer as a starting point for editing into a final version? Please also provide a brief justification for your preference in the `Notes` column.\\

WARNING: Unfortunately, it is not guaranteed that the draft letters accurately reflect the referral reason or patient history. This will need to be ascertained based on the provided medical note and should heavily factor into your preference.\\
\\
\textbf{Options:} \\
- A Strongly Preferred - A Somewhat Preferred - Tied - B Somewhat Preferred - B Strongly Preferred\\
\end{tcolorbox}

Three different U.S. board certified physicians are recruited and each of them evaluates all 25 examples. Ratings are aggregated by mapping the Likert scales to a numerical range ([-2,2]) and taking the sign of the median value.

\paragraph{Medical simplification evaluation}
This task involves generating a plain language summary (PLS) from a technical abstract from a biomedical systematic review. A PLS is a version of the technical abstract that is written in plain English and meant to be understood by most readers without a university education \citep{cochrane-pls}. 

A set of 25 technical abstracts and plain language summaries from systematic reviews conducted by Cochrane is sampled from the test split of the dataset introduced by \cite{devaraj-etal-2021-paragraph}. The expert plain language summaries are written by the original authors of the Cochrane systematic reviews.

Med-Gemini-M 1.0 is prompted to generate a PLS given the technical abstract as follows:
\begin{tcolorbox} \small

Please read through the provided technical summary of a body of medical research and provide a simplified summary that is accessible to a lay audience without medical expertise.\\\\

Technical Summary: \{TECHNICAL\_ABSTRACT\}\\
Simplified Summary:

\end{tcolorbox}

Clinicians are presented with the technical abstract, the original PLS and our model's generated PLS. They are blinded to the source of each PLS and asked to perform the following comparisons:

\begin{tcolorbox} \small
\textbf{Grounding:} \\
Is all the information in the simple summary factually supported by the technical summary?\\
- A Strongly Preferred - A Somewhat Preferred - Tied - B Somewhat Preferred - B Strongly Preferred\\
\\
\textbf{Coverage:} \\
Are the most important takeaways for a lay audience included in the simple summary?\\
- A Strongly Preferred - A Somewhat Preferred - Tied - B Somewhat Preferred - B Strongly Preferred\\

\textbf{Succinctness:} \\
Does the simple summary only contain the most important takeaways for a lay audience?\\
- A Strongly Preferred - A Somewhat Preferred - Tied - B Somewhat Preferred - B Strongly Preferred\\

\textbf{Reading Level:} \\
Is the reading-level of the simple summary appropriate for a lay audience?\\
- A Strongly Preferred - A Somewhat Preferred - Tied - B Somewhat Preferred - B Strongly Preferred\\

\textbf{Overall:} \\
What is the overall quality of the simple summary for a lay audience?\\
- A Strongly Preferred - A Somewhat Preferred - Tied - B Somewhat Preferred - B Strongly Preferred\\
\end{tcolorbox}

Three different U.S. board certified physicians each evaluates all 25 examples. Ratings are aggregated similar to the referral letter task.

\setcounter{table}{0}
\setcounter{figure}{0}
\renewcommand{\thetable}{D\arabic{table}}
\renewcommand{\thefigure}{D\arabic{figure}}
\section{Additional details on multimodal understanding tasks} 

\subsection{Multimodal fine-tuning datasets}\label{app:mm_ft}
\begin{figure}[ht!]
    \centering
    \includegraphics[width=\textwidth]{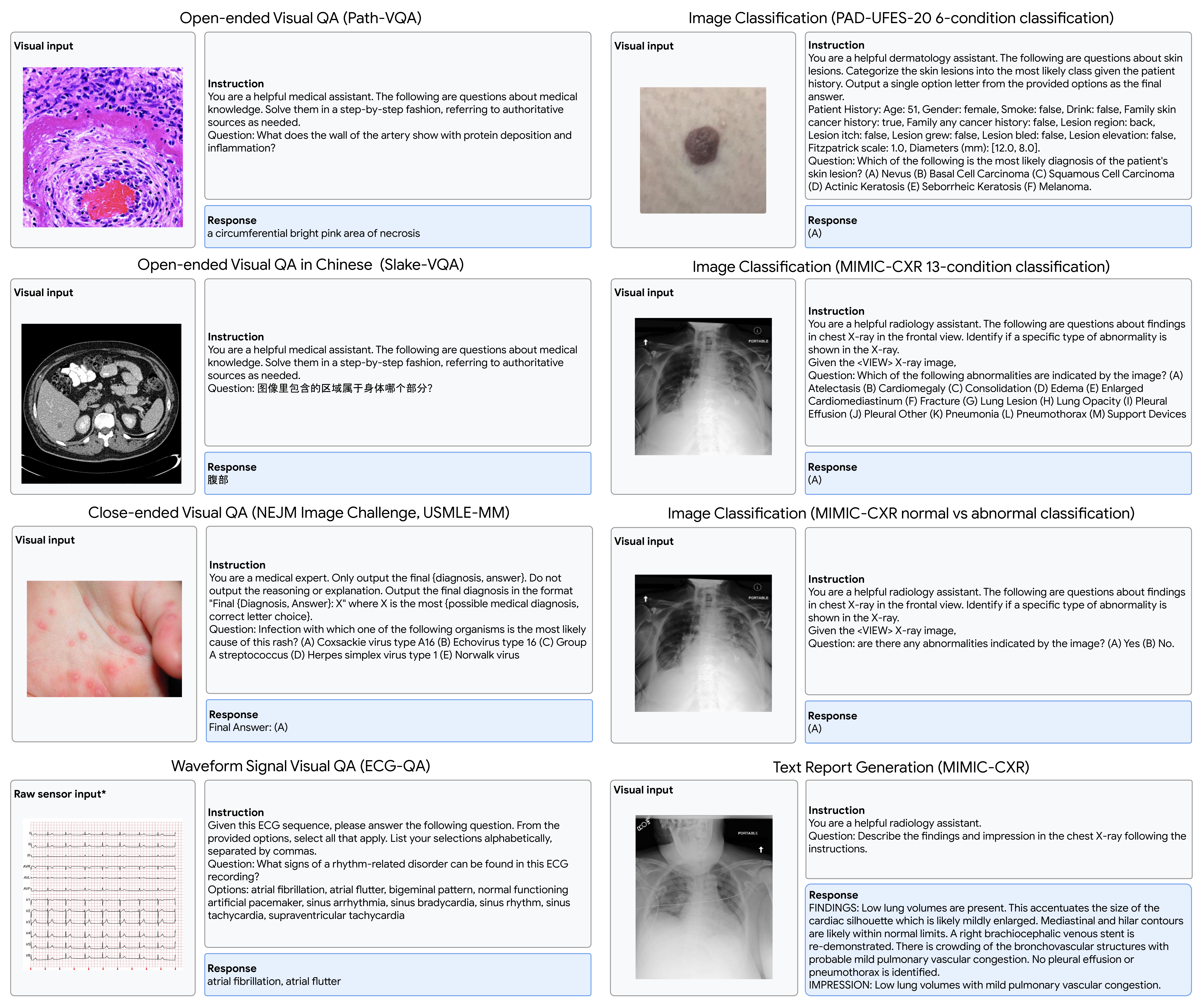}
    \caption{\footnotesize\textbf{Representative examples and prompts of multimodal understanding tasks.} Med-Gemini is evaluated on various tasks, including image classification and visual question answering (VQA), demonstrating its ability to analyze and interpret diverse biomedical data formats. Note that the input for ECG-QA is the raw ECG sensor sequence, visualized here as a 12-lead ECG image from PTB-XL~\citep{wagner2020ptb}. Also note that three MIMIC-CXR tasks are only used for instruction fine-tuning.}
    \label{fig:mm_ex}
\end{figure}

For Med-Gemini-M 1.5's multimodal fine-tuning, we use four image-text datasets from MultiMedBench~\citep{tu2024towardsmpm, tanno2024consensus} including Slake-VQA~\citep{liu2021slake}, Path-VQA~\citep{he2020pathvqa}, MIMIC-CXR~\citep{johnson2019mimic,johnson2019mimicjpg}, PAD-UFES-20~\citep{pacheco2020pad}, in addition to the Radiology Objects in COntext (ROCO) dataset~\citep{pelka2018radiology}. We further use a subset of ECG-QA~\citep{oh2023ecgqa} to develop the health signal encoder for encoding sensor input in Med-Gemini-S 1.0. We describe the datasets in details below:
\begin{table}[ht!]
\footnotesize
\centering
\resizebox{1.0\textwidth}{!}{
\begin{tabular}{cccccc}
\toprule
\textbf{Task type} & \textbf{Dataset} & \textbf{\begin{tabular}[c]{@{}c@{}}Fine-tuning \\ sample size\end{tabular}} & \textbf{Dataset Description} & \textbf{Reference} \\
\midrule
VQA & Slake-VQA & 9849 & Close/open-ended English/Chinese VQA of radiology images & \cite{liu2021slake} \\
VQA & Path-VQA & 19755 & Close/open-ended VQA of pathology images & \cite{he2020pathvqa} \\
VQA & ROCO & 29907 & Close/open-ended VQA of radiology and non-radiology images & \cite{pelka2018radiology} \\
\midrule
Classification & PAD-UFES-20 & 1838 & Close-ended, multiple-choice, 6-class dermatology condition & \cite{pacheco2020pad} \\
Classification & MIMIC-CXR & 164512 & Close-ended, multiple-choice, 13-class CXR condition & \cite{johnson2019mimic,johnson2019mimicjpg} \\
Classification & MIMIC-CXR & 237962 & Close-ended, binary-choice, normal vs. abnormal classification & \cite{johnson2019mimic,johnson2019mimicjpg} \\
\midrule
Text Report Generation & MIMIC-CXR & 90968 & Open-ended, predicting CXR finding given image and indication & \cite{johnson2019mimic,johnson2019mimicjpg} \\
\midrule
Signal QA & ECG-QA & 159306 & Close-ended signal QA of electrocardiograms. & \cite{oh2023ecgqa} \\
\bottomrule
\end{tabular}
}
\caption{\footnotesize\textbf{Overview of the datasets used for multimodal instruction fine-tuning.}}
\label{tab:mm_ft}
\end{table}
\begin{itemize}
    \item \textbf{MIMIC-CXR} is a CXR dataset with free-text reports~\citep{johnson2019mimic,johnson2019mimicjpg}, consisting of 377110 chest X-ray images along with the corresponding protected health information (PHI)-removed text reports from 65379 patients (227835 image studies, with one or more image view positions). Each report is annotated with 13 common radiological conditions using the CheXpert labelling software ~\citep{irvin2019chexpert}. We use the official train/test split as described in the MIMIC-CXR for all tasks. We consider four fine-tuning tasks using MIMIC-CXR: (1) normal vs. abnormal binary classification, (2) CXR abnormality condition VQA, (3) synthetic CXR VQA, and (4) text report generation. For the normal vs. abnormal binary classification task, we classify each image into either normal or abnoraml category based on the CheXpert ``no finding'' label using all frontal view images [anterior-posterior (AP) and posterioranterior (PA) views] with the task prompt listed in~\Cref{fig:mm_ex}. For CXR abnormality condition VQA, we exclude all images with normal findings, and group positive and uncertain labels as positive class for 13 abnormal conditions: atelectasis, cardiomegaly, consolidation, edema, enlarged cardiomediastinum, fracture, lung lesion, lung opacity, pleural effusion, pleural other, pneumonia, pneumothorax, and support devices. Then we frame the abnormality detection problem into a close-ended multi-class multiple-choice question setup as shown in~\Cref{fig:mm_ex}. To further enrich these VQA tasks, we generate an collection of synthetic question-and-answer pairs from radiology reports by querying Gemini base models. We specifically prompt the LLM to extract pairs of yes-or-no question and the corresponding answer from each report such that they are independent of the presence of the above 13 conditions. We ensure that for each question, the number of ``yes'' and ``no'' are the same to avoid introducing spurious correlation. All VQA tasks are added as the auxiliary tasks for the report generation task which combines the image with the contextual information from the INDICATION section (reason for the study) as the model input to generate the FINDINGS and IMPRESSION sections of the report as the target, similar to prior works~\citep{hyland2023maira, tu2024towardsmpm}. Furthermore, following the procedure proposed in~\cite{tanno2024consensus}, we filter out the training examples whose reports reference prior studies and only keep examples where the report only refers to findings present in the input image. This aims to mitigate hallucination of references to non-existing prior reports, a common issue raised by multiple lines of research \cite{pmlr-v193-ramesh22a} and \cite{hyland2023maira}. 
    The evaluation of MIMIC-CXR will be reported in a subsequent paper.
    \item \textbf{PAD-UFES-20} includes 2298 clinical skin lesion images collected from various smartphone devices with different resolutions, sizes, and lighting conditions through the Dermatological and Surgical Assistance Program at the Federal University of Espírito Santo (UFES-Brazil)~\citep{pacheco2020pad}. Six types of skin lesions are included in the dataset: basal cell carcinoma, melanoma, squamous cell carcinoma, actinic keratosis, melanocytic nevus, and seborrheic keratosis. Each image is correlated with up to 21 clinical features (e.g., patient demographics, family cancer history lesion location, lesion size).
    Given no published official splits, we adopt two PAD-UFES-20 split setup. We use Med-PaLM M split (the image-level split) for a direct, fair evaluation and comparison against the previous SoTA method. We also evaluate on a new split, which is a split at the patient level (\Cref{tab:mm_perf}).
    We set up three classification tasks for fine-tuning: (1) 6-class classification using the original label distribution and 14 clinical features (age, gender, smoke, drink, skin cancer history, cancer history, region, Fitspatrick, horizontal and vertical diameters, itch, grew, bleed, and elevation); (2) 6-class classification using images and clinical features as the previous task, but with image augmentation on the training set using 8 RandAugment~\citep{cubuk2020randaugment} operations: autoContrast, equalize, invert, rotate, posterize, solarize, color, and contrast; (3) 6-class classification the same as previous task, but using an upsampled subset for four minor skin conditions (melanoma, squamous cell carcinoma, seborrheic keratosis, and nevus) with image augmentation during training to mitigate the class imbalance problem.
    The latter two auxiliary tasks are included in the training mixture to help the model to distinguish among different types of clinical observations.
    We also formulate the skin condition classification problem as a close-ended multiple-choice question setup as shown in~\Cref{fig:mm_ex}, and report the prediction accuracy for this task.
    \item \textbf{Path-VQA} is a pathology VQA dataset, which consists of 998 pathology images with 32799 QA pairs~\citep{he2020pathvqa}. All images are extracted from medical textbooks and online digital libraries. Each image is associated one or more questions regarding different aspects of the pathology imaging including color, location, appearance, shape, etc. 50.2\% of the QA pairs are open-ended questions (divided into 7 categories: what, where, when, whose, how, and how much/how many). 49.8\% of the QA pairs are close-ended questions with simple "yes/no" answer. We adopt the official splits where the training/validation/testing splits contain 19755, 6279, and 6761 QA pairs, respectively.
    \item \textbf{Slake-VQA} is a bilingual (English and Chinese) radiology image VQA dataset~\citep{liu2021slake}, containing 642 annotated images with 14028 question-answer pairs covering three imaging modalities (CT, MRI, and chest X-Rays), 39 organ systems, and 12 diseases. Questions are either open-ended or closed-ended related to various aspects of the radiology images, including plane, quality, position, organ, abnormality, size, color, shape, knowledge graph, etc. The training/validation/testing splits contain 9849, 2109, and 2070 QA pairs, respectively.
    \item \textbf{ROCO} (Radiology Objects in Context) dataset is a large-scale medical and multimodal imaging dataset~\citep{pelka2018radiology}. The ROCO images are from publications available on the PubMed Central Open Access FTP mirror, which are automatically labeled as either radiology or non-radiology. Each image has its caption, keywords, the corresponding UMLS Semantic Types (SemTypes), and UMLS Concept Unique Identifiers (CUIs). We use the official training set across radiology and non-radiology, which contain 29907 image-caption pairs, and set up a captioning task for fine-tuning. We only include the images under CC BY, CC BY ND, CC BY SA and CC0 licenses in ROCO.
    \item \textbf{ECG-QA} is a sensor-text multimodal benchmark for assessing cardiac health~\citep{oh2023ecgqa}. It is the first QA dataset specifically designed for electrocardiogram analysis based on PTB-XL~\citep{wagner2020ptb}, containing diverse question templates, each validated by an ECG expert to ensure clinical utility. Strong performance on ECG-QA indicates the ability to grasp complex medical concepts and their connections to raw waveform signals. ECG-QA contains two types of questions involving (1) single ECG and (2) comparing two ECGs; each question type consists of (1) yes/no questions, (2) multiple-choice questions, and (3) open-ended questions to provide ECG-related attributes. We focus on single ECG questions in this work, which contain 159306, 31137, and 41093 samples for train, validation and test sets, respectively.

\end{itemize}

\subsection{Multimodal evaluation datasets}\label{app:mm_eval}
In addition to the in-distribution datasets (details in the above section), we include three out-of-distribution datasets to evaluate the mulitmodal capability of Gemini:

\begin{table}[ht!]
\footnotesize
\centering
\resizebox{1.0\textwidth}{!}{
\begin{tabular}{cccccc}
\toprule
\textbf{Task Type} & \textbf{Modality} & \textbf{Dataset} & \textbf{Test sample size} & \textbf{Description} & \textbf{Reference} \\
\midrule
\multirow{7}{*}{VQA} & Radiology & Slake-VQA & 2070 & English-Chinese bilingual VQA on radiology images & \cite{liu2021slake} \\
 & Pathology & Path-VQA & 6761 & Close/open-ended VQA on pathology images & \cite{he2020pathvqa} \\
 & Dermatology & PAD-UFES-20 & 460 & 6-class skin condition multiple-choice & \cite{pacheco2020pad} \\
 & Cross-specialty & NEJM Image Challenge & 934 & OOD, Close-ended VQA on open domain medical images & \cite{nejm_image_challenge} \\
 & Cross-specialty & USMLE-MM & 46 & OOD, Close-ended VQA on open domain medical images & Novel \\
 & Cross-domain & MMMU-HM & 150 & OOD, Close/open-ended VQA on health and medical images & \cite{yue2023mmmu} \\
\midrule
Signal QA & Cardiology & ECG-QA & 41093 & Close-ended signal QA of electrocardiograms & \cite{oh2023ecgqa} \\
\bottomrule
\end{tabular}
}
\caption{\footnotesize\textbf{Overview of the datasets used for multimodal understanding evaluation}. OOD: out-of-distribution dataset.}
\label{tab:mm_eval}
\end{table}

\begin{itemize}
    \item \textbf{New England Journal of Medicine (NEJM) Image Challenge} is a renowned clinical case challenge series that tests the diagnostic acumen and visual observation skills of medical professionals worldwide \citep{nejm_image_challenge}. Every week, the NEJM presents a clinical image accompanied by a brief case description. The images include radiographic images, natural and dermatoscopic skin images, electrocardiograms, histopathology images, endoscopy images, and ophthalmoscopy images. Readers are invited to carefully analyze the photograph, consider the patient's history, and select the final diagnosis from five possible diagnosis candidates. We collect 942 NEJM Image Challenge cases from 2005 to 2023. Each case consists of a medical image and an associated question (e.g., ``What is the most likely diagnosis?''), five multiple-choice options, and a correct answer. Some cases additionally provide text captions with relevant clinical context or other background information in the question. We have collected 942 cases in total, yet 934 cases are evaluated in the end for the fair comparison (until October 12, and two cases, 20160519 and 20111103 were not evaluated due to GPT-4V filters preventing images that are assumed to be sexually explicit~\citep{buckley2023accuracy}).
    \item \textbf{USMLE-MM (Multimodal)} is a multimodal multiple-choice question dataset with 46 questions identified in the sample exams provided by \texttt{www.usmle.org}, which includes images in the question. The sample exams are used for USMLE preparation.
    \item \textbf{MMMU-HM (health and medicine)} is a subset of the publicly available benchmark, MMMU (Massive Multi-discipline Multimodal Understanding) validation set~\citep{yue2023mmmu}. MMMU-HM includes 150 questions related to basic medical science, clinical medicine, diagnostics and laboratory medicine, pharmacy, and public health domains.
\end{itemize}

\subsection{Additional results for multimodal tasks}
\paragraph{ECG-QA} To expand Med-Gemini's capability to process raw biomedical signals for ECG-QA tasks, we augment Gemini 1.0 Nano with an ECG-specific encoder and fine-tune using two approaches: keeping Gemini model unchanged (\textit{frozen}) and fine-tuning Gemini model (\textit{unfrozen}). We compare our Med-Gemini-S 1.0 to their baseline counterparts: our model with \textit{frozen} Gemini model to GPT-4 with SE-WRN ECG features in input prompts~\citep{oh2023ecgqa} and our model with \textit{unfrozen} Gemini model to an ECG foundation model based on M\textsuperscript{3}AE~\citep{oh2023ecgqa}. Med-Gemini-S 1.0 with \textit{frozen} and \textit{unfrozen} Gemini yield accuracies of 57.7\% and 58.4\% on single ECG questions, respectively, outperforming GPT-4 (51.6\%) by 6.1\% and M\textsuperscript{3}AE (57.6\%) by 0.8\%.

\setcounter{table}{0}
\setcounter{figure}{0}
\renewcommand{\thetable}{E\arabic{table}}
\renewcommand{\thefigure}{E\arabic{figure}}
\section{Additional details on long-context understanding tasks}

\subsection{Long-context evaluation datasets}\label{app:lc_eval}
\begin{itemize}
    \item \textbf{MIMIC-III Needle-in-a-Haystack} is a specially curated dataset from MIMIC-III~\citep{johnson2016mimic} for subtle medical condition search-retrieval task over long EHRs. It is designed to mimic a clinically-relevant ``needle-in-a-haystack'' challenging problem~\citep{team2023gemini}. MIMIC-III is a large publicly-available medical database that contains medical records of patients admitted to intensive care units. We randomly select unstructured medical notes from 44 unique patients with more than 100 ``high-value''\footnote{Consult notes, Progress notes, History and Physical notes and Discharge Summary notes authored by physicians/PAs/NPs/APRNs and Operative notes by physicians/PAs.} clinical notes. To construct ``needle-in-a-haystack'' examples, we use our prior work~\citep{feder2022building}, which aims at identifying the problem list (conditions/symptoms/procedures) from patients' collection of EHR documents through (1) labeling all mentions (text spans) of problems on the medical records with machine learning based annotators; (2) rule-based selection and aggregation of mentions to decide whether a problem is actually existent or not. We select the examples where there is only 1 evidence snippet identified in the aggregation step, and then randomly sample 100 negative and 100 positive examples determined by the rule-based method. 200 selected examples are then sent to 3 human medical raters to decide whether the problem actually exists or not. Specifically, the raters are presented with the condition name and retrieved supporting evidence snippets. The raters are then asked to answer the question: ``Select ALL problems the patient HAS HAD based on the evidence in the provided note excerpts.'' As a result, we have 121 positive examples and 79 negative examples based on the majority voting\footnote{Example of a negative example: a patient's records with one mention of \textit{Sepsis} in a text segment ``Received IV Ceftriaxone for concern of UTI/sepsis.''. Here the patient should not be labeled as having history of sepsis as there is no definitive diagnosis of the condition without other context.} (Krippendorff's alpha at 0.77, see~\Cref{tab:app_ehr_rater}). The majority-vote labels are then used as ground truth labels for the subsequent evaluation. For each example, it consists of a set of medical records, a test question regarding to whether or not a condition of interest exists, and a binary ground truth label. The length of the medical records varies from 200K to 700K words.
    \item \textbf{Medical Instructional Video QA (MedVidQA)} is a video-language cross-modal dataset for the Medical Visual Answer Localization (MVAL) task~\citep{gupta2023dataset}. Three medical informatics experts created 3010 health-related instructional questions for 899 videos extracted from YouTube, and localized the visual answer to those questions by annotating their timestamps in the video, i.e., identifying the timestamp span given text question query. The mean duration time of these videos is 383.29 seconds. We follow the official data split, where 2710, 145, and 155 questions and visual answers are used for training, validation, and testing respectively. However, 7 questions are excluded due to the YouTube video access restriction (private videos, removed videos).
    \item \textbf{Cholec80} and \textbf{Cholec80-CVS}. Cholec80 is a dataset containing 80 high-quality videos of laparoscopic cholecystectomy performed by 13 surgeons~\citep{twinanda2016endonet}. Cholec80 is one of the most popular benchmarks for research in laparoscopic cholecystectomy video analysis with deep learning, and it has been widely used in recent research, on different video understanding tasks, including temporal segmentation of surgical phases~\citep{golany2022artificial, chen2018endo3d}, and surgical tool detection~\citep{nwoye2019weakly, leifman2022pixel}. Cholec80-CVS~\citep{rios2023cholec80} contains Critical View of Safety (CVS) criteria annotations, provided by skilled surgeons, for each video in the Cholec80 dataset.  The CVS~\citep{strasberg2010rationale} is a mandatory method, defined by three visual criteria, used for secure identification of the cystic duct and cystic artery to minimize the risk of Bile Duct Injury (BDI).
    For each video in Cholec80, skilled surgeons selected different video segments where at least one CVS criteria was satisfied, and then for each selected video segment, the surgeons assigned a score of 0, 1, or 2 for each of the three CVS criteria, following an extension of the original scoring system proposed by~\citep{sanford2014simple} and~\citep{mascagni2021surgical}. In total, Cholec80-CVS provides CVS criteria annotations for 572 video segments within Cholec80 videos.
    We assess the performance of Med-Gemini-M 1.5 in comparison to GPT-4V and Resnet3D. It is important to note that GPT-4V does not officially support video data as input. Therefore, we sample frames from each video clip at a rate of 1 frame per second and combine a sequence of frames as input to the model. During our experimentation, we observe that GPT-4V's vision context length is limited, and we are able to insert up to 300 low-resolution images. Consequently, we filter out all video clips longer than 5 minutes. For fair comparison, we evaluate Med-Gemini-M 1.5 on the same filtered subset of video clips. To conduct the evaluation on Resnet3D, we randomly split the dataset into 5 consecutive folds and assess the performance on each validation fold separately. The average accuracy across all five folds is reported.

\end{itemize}

\begin{table}[ht!]
\footnotesize
\centering
\resizebox{1.0\textwidth}{!}{
\begin{tabular}{cccccc}
\toprule
\textbf{Task Type} & \textbf{Modality} & \textbf{Dataset} & \textbf{Test sample size} & \textbf{Description} & \textbf{Reference} \\
\midrule
EHR & Text & MIMIC-III-Needle-in-a-Haystack & 200 & Problem identification from EHR records  & Curated based on~\cite{feder2022building} \\
\midrule
MVAL & Video $\pm$ Text & MedVidQA & 148 & Video span prediction & \cite{gupta2023dataset} \\
\midrule
Classification & Video & Cholec80/Cholec80-CVS & 572 & Critical view of safety assessment & \cite{rios2023cholec80} \\
\bottomrule
\end{tabular}
}
\caption{\footnotesize\textbf{Overview of the datasets used for the long-context capability evaluation.} MVAL: medical visual answer localization.}
\label{tab:lc_eval}
\end{table}

\begin{table}[ht!]
\centering
\resizebox{0.8\textwidth}{!}{
\begin{tabular}{cccc}
\toprule
\textbf{Methods} & \textbf{Precision} & \textbf{Recall} & \textbf{F1} \\
\midrule
Heuristic-based baseline & \textbf{0.85 (0.78, 0.92)} & 0.73 (0.64, 0.80) & \textbf{0.78 (0.72, 0.84)} \\
Med-Gemini-M 1.5 (one-shot) & 0.77 (0.66, 0.86) & \textbf{0.76 (0.67, 0.86)} & 0.77 (0.68, 0.84) \\
\bottomrule
\end{tabular}
}
\caption{\footnotesize\textbf{Performance comparison of Med-Gemini-M 1.5 versus the heuristic-based annotation-aggregation baseline.}}
\label{tab:lc_ehr}
\end{table}

\subsection{Rater agreement metrics for the long EHR understanding task}

To ensure the reliability of the EHR benchmark, we collect ratings from three independent raters for each of the 200 example questions.  The following metrics demonstrate strong consistency among raters:

\begin{itemize}
    \item Jaccard Similarity Index: Measures the overlap between sets of rater selections. Let $A$, $B$, and $C$ represent the sets of selections made by each rater. The Jaccard similarity index for unanimous selections is defined as $J_{=3} = \frac{|A \cap B \cap C|}{|A \cup B \cup C|}$. The Jaccard similarity index for at least 2 raters being in agreement is defined as $J_{\geq2} = \frac{|(A \cap B) \cup (A \cap C) \cup (B \cap C)|}{|A \cup B \cup C|}$.
    \item Krippendorff's Alpha:  A reliability coefficient designed for multiple raters. 
\end{itemize}

\begin{table}[ht!]
\centering
\resizebox{0.8\textwidth}{!}{
\begin{tabular}{ccccc}
\toprule
\textbf{} & \textbf{Num. tasks} & \textbf{$J_{=3}$} & \textbf{$J_{\geq2}$} & \textbf{Krippendorff’s alpha} \\
\midrule
Existence of condition & 200 & 0.83 & 0.915 & 0.77 \\
\bottomrule
\end{tabular}
}
\caption{\footnotesize\textbf{Rater agreement metrics on long EHR understanding task.}}
\label{tab:app_ehr_rater}
\end{table}

A Jaccard similarity index of $0.83$ for unanimous selections indicates substantial agreement when all three raters select identical choices.
An even higher Jaccard index of $0.915$ reflects strong consistency when at least two out of three raters make the same selections.
A Krippendorff’s alpha of $0.77$ indicates good agreement on the existence of medical conditions within the EHR data.

%% file: related.tex
\section{Related Works}
\paragraph{Overview of large language model in medicine}
Large language models (LLMs) have revolutionized machine learning and artificial intelligence. Researchers have employed novel network architectures, such as transformers~\citep{vaswani2017attention} and pathways~\citep{barham2022pathways}, to train these models on massive datasets. This self-supervised training across diverse domains includes models like BERT~\citep{devlin2018bert}, GPT~\citep{radford2018improving}, T5~\citep{raffel2020exploring}, FLAN~\citep{wei2021finetuned}, BLOOM~\citep{le2022bloom},  Flamingo~\citep{alayrac2022flamingo}, PaLM and PaLM2~\citep{chowdhery2023palm,anil2023palm}, LLaMA~\citep{touvron2023llama}, PaLI~\citep{chen2022pali}, PaLM-E~\citep{driess2023palm}, and the recent Gemini models~\citep{team2023gemini,team2024gemini}. By processing  text or multimodal information, these pretrained models develop a robust understanding of language, patterns, and relationships with remarkable adaptibility.

Minimal fine-tuning allows these models to adapt to diverse downstream tasks. In the medical domain, Med-PaLM~\citep{singhal2023large} and Med-PaLM 2~\citep{singhal2023towards}  represent pioneering medical LLMs fine-tuned on EHRs, exam questions, and research literature. To achieve the goal of generalist medical AI (GMAI)~\cite{moor2023foundation}, researchers use general LLMs with prompting strategies [e.g., GPT-4 with Medprompt~\citep{nori2023can}], or refine them with multimodal data for enhanced medical understanding [e.g., Med-PaLM-M~\citep{tu2024towardsmpm}]. These models show promise in diagnosis assistance~\citep{mcduff2023towards}, risk prediction, drug discovery, diagnostic dialogue~\citep{tu2024towardsamie} and assessing psychiatric functioning~\citep{galatzer2023capability}. Our work leverages the latest Gemini models, using either direct instruction prompting or further fine-tuning for specialized medical tasks. Below, we discuss related works across the areas of language, multimodal learning, and long-context modeling.

\paragraph{Model reasoning and tool-use for language-based tasks}
Reasoning is a process of logical thinking that leads to a conclusion, which can be significantly enhanced by recent advances in LLMs and large multimodal models (LMMs). These improvements stem from a combination of better models and methods that directly imitate human reasoning. Language model based reasoning techniques have been surveyed in prior works~\citep{qiao2023reasoning, huang2023reasoning}, with such surveys extended into multimodal reasoning~\citep{wang2024exploring}. Strategies to enhance language reasoning include prompt engineering, improved processes, and enhancing reasoning with access to external elements such as tools or knowledge. Prompt engineering is exemplified by approaches such as Chain-of-Thought (CoT) prompting \citep{wei2022chain}, which involves generating a series of intermediate reasoning steps, Least-to-Most prompting, which involves breaking down a problem into smaller subproblems and then sequentially solving them \citep{zhou2023leasttomost}, and other methods that explore different reasoning paths to arrive at a conclusion \citep{yao2023tree, besta2024graph}. Improved processes arise from methods such as model updates via self-improvement \citep{zelikman2022star} or ensemble-based approaches \citep{wang2022self}. 

Access to external elements such as tools~\citep{schick2024toolformer,hao2024toolkengpt} or external knowledge bases through the use of retrieval augmented generation (RAG)~\citep{gao2024retrievalaugmented,zhang2024raft} has also demonstrated improvements in language model reasoning. Recently LLMs have also evolved to interact with information and web tools. For tool-use, LLMs can learn to execute external tools or application programming interface (APIs), enabling them to perform actions in the real world like searching, calendar use, or using translation service via APIs~\citep{schick2024toolformer,qin2023toolllm}. For web search specifically, LLMs incorporate traditional search engines by understanding complex queries and providing summaries that synthesize information from multiple sources~\citep{nakano2021webgpt, varshney2023stitch}. Furthermore, LLMs are able to not only retrieve information but also utilize tools and create ones based on user-defined needs~\citep{cai2023large}. \cite{zakka2024almanac} have demonstrated that search tool-use can be particularly useful in medical guideline and treatment recommendations. In this work, we integrate a strategy of self-training with search to improve Med-Gemini's capabilities for model reasoning.

\paragraph{Large multimodal models in medicine}
Medical practice often requires integration of multiple modalities to deliver effective care, for example, integrating data sources from patient history, medical imaging, genetic testing and lab results. Models that can integrate such modalities may provide a more comprehensive picture of a patient's condition. Existing approaches fall into two broad categories: specialist and generalist. Specialist models excel at specific tasks within a medical discipline. Examples include models optimized for radiology report generation \citep{tanno2024consensus, zambranochaves2024training}, pathology question answering or histopathology captioning \citep{lu2023foundational}, radiology-related tasks \citep{xu2023elixr}, and cardiology electrocardiogram captioning \citep{wan2024electrocardiogram}. 

Conversely, ``generalist medical AI'' (GMAI) systems ~\citep{moor2023foundation}, such as Med-PaLM M \citep{tu2024towardsmpm} and LLaVA-Med \citep{li2024llava}, tackle a wider range of tasks across multiple specialties, aiming for broader applicability in clinical settings. The diversity of tasks performed by systems such as Med-PaLM M performance remains noteworthy as one of the earliest examples of generalist multimodal models in medicine, capable of addressing  radiology, pathology, dermatology, and genomics tasks with competitive performance or exceeding SoTA across different specialties using a strong pretrained LLM with appropriate fine-tuning strategies. In this report, we further advance the evidence that AI systems can deliver strong generalist multimodal capabilities in medicine with Med-Gemini but the primary focus is on developing a model family considering application specific trade-offs.

\paragraph{Long-context capability of large language models}
Prior works addressing tasks with long-context windows have been limited by the capabilities of LLMs to effectively utilize large spans of text due to the memory and computation limitation of the Transformer-based models~\citep{liu2024lost,vaswani2017attention}. Initial efforts used hierarchical approaches to derive representations of clinical text that could not fit into a model's limited context window \citep{dai2022revisiting}. Subsequent work such as Clinical-Longformer and Clinical-BigBird \citep{li2023comparative} focused on extending context lengths from 512 to 4096 tokens, enabling improvements in performance in question answering, document classification and information retrieval tasks. Subsequent approaches explored the use of such models in combination with imaging encoders to tackle multimodal tasks such as medical visual question answering \citep{gupta-demner-fushman-2022-overview}. With the advancements of hardware and efficient algorithms, researchers have developed LLMs with larger context windows toward 100K tokens~\citep{dai2019transformer,poli2023hyena}. Recently, Gemini further advanced the boundary of long-context capability to one million tokens~\citep{team2024gemini}.

However, in the domain of medicine the majority of LLMs continue to be evaluated on relatively short texts \citep{parmar2023longbox} and single images. Despite their importance to medicine and clinical practice, long-context capabilities in medicine, especially in multimodal settings, are underexplored. We address this unmet need and investigate the potential of Med-Gemini on different long-context use cases, including video and long EHR-related tasks.